\documentclass[5p,twocolumn]{elsarticle}

\usepackage{hyperref}
\usepackage{booktabs}
\usepackage{amsfonts}       % blackboard math symbols
\usepackage{nicefrac}       % compact symbols for 1/2, etc.
\usepackage{amsmath}
\usepackage{multirow}
\usepackage{longtable}
\usepackage{color}

\newcommand{\difftxt}[1]{\textcolor{black}{#1}}
\newcommand{\LT}[1]{\textcolor{black}{#1}}

\begin{document}

\begin{frontmatter}

\title{Multiple Hypothesis Colorization and Its Application to Image Compression}

\author[mymainaddress]{Mohammad Haris Baig\corref{mycorrespondingauthor}}
\cortext[mycorrespondingauthor]{Corresponding author}
\ead{haris@cs.dartmouth.edu}

\author[mymainaddress]{Lorenzo Torresani}
\ead{LT@dartmouth.edu}

\address[mymainaddress]{Hanover, New Hampshire. United States}

\begin{abstract}
In this work we focus on the problem of colorization for image compression. Since color information occupies a large proportion of the total storage size of an image, a method that can predict accurate color from its grayscale version can produce a dramatic reduction in image file size. But colorization for compression poses several challenges. First, while colorization for artistic purposes simply involves predicting plausible chroma, colorization for compression requires generating output colors that are as close as possible to the ground truth. Second,  many objects in the real world exhibit multiple possible colors. Thus, in order to disambiguate the colorization problem some additional information must be stored to reproduce the true colors with good accuracy. 
To account for the multimodal color distribution of objects we propose a deep tree-structured network that generates for every pixel multiple color hypotheses, as opposed to a single color produced by most prior colorization approaches. We show how to leverage the multimodal output of our model to 
reproduce with high fidelity the true colors of an image by storing very little additional information. In the experiments we show that our proposed method outperforms traditional JPEG color coding by a large margin, producing colors that are nearly indistinguishable from the ground truth at the storage cost of just a few hundred bytes for high-resolution pictures! 
\end{abstract}

%\texttt{elsarticle.cls}\sep 
\end{frontmatter}

%\linenumbers

\section{Introduction}

\begin{table*}[t!]\label{tb:teaser}
\centering
\setlength{\tabcolsep}{1pt}
\begin{tabular}{c|c|ccc}
\toprule
 Image & JPEG~\cite{jpeg} & \multicolumn{3}{|c}{Our approach} \\ 
\hline
 & $1,207$ bytes & $192$ bytes & $338$ bytes & $531$ bytes\\
\includegraphics[width=3.0cm]{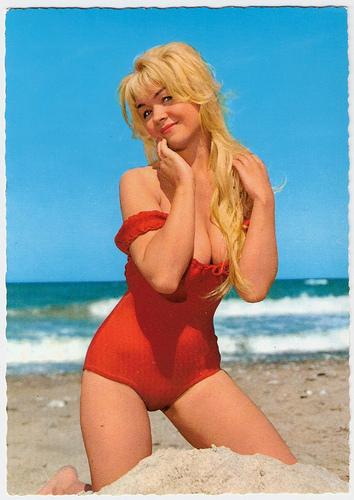}&\includegraphics[width=3.0cm]{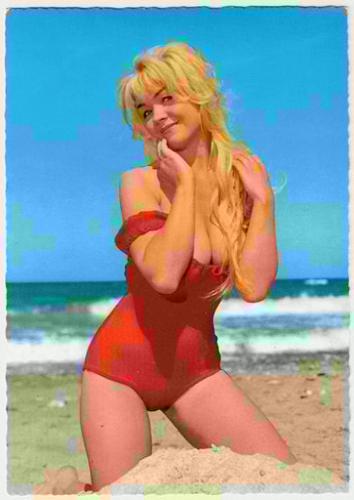}&\includegraphics[width=3.0cm]{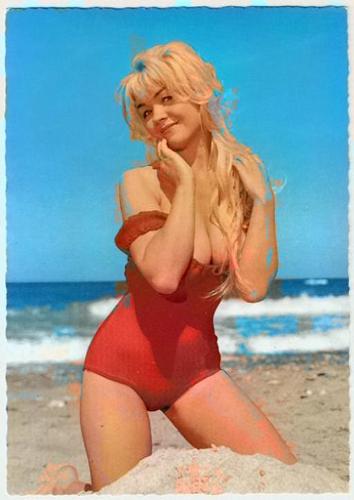}&\includegraphics[width=3.0cm]{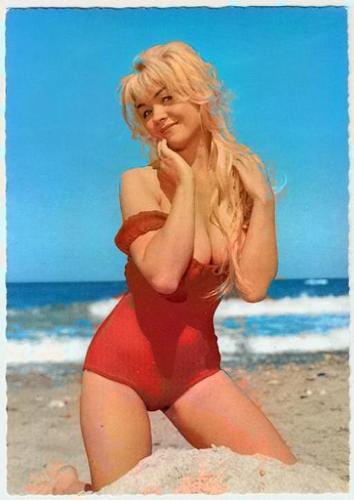}&\includegraphics[width=3.0cm]{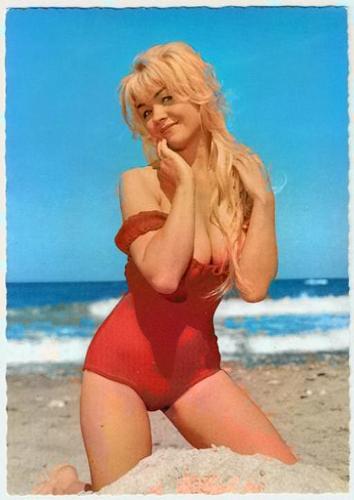}\\ 
\hline
 & $1,104$ bytes & $198$ bytes & $383$ bytes & $622$ bytes\\
\includegraphics[width=3.0cm]{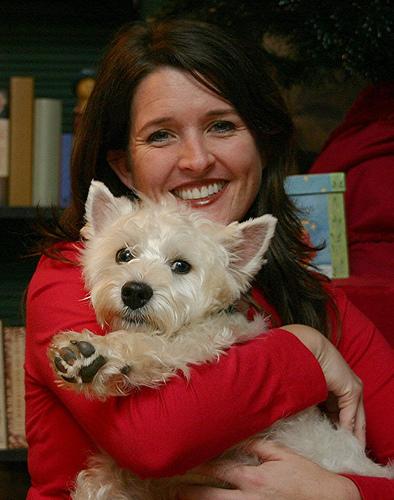}&\includegraphics[width=3.0cm]{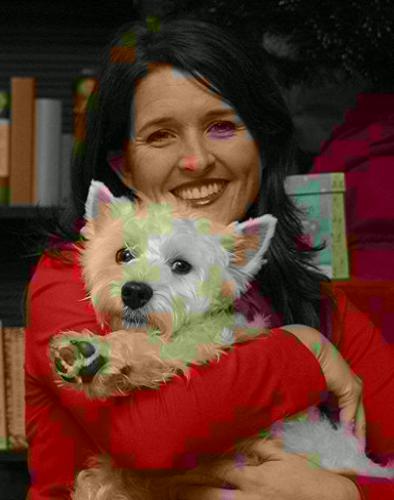}&\includegraphics[width=3.0cm]{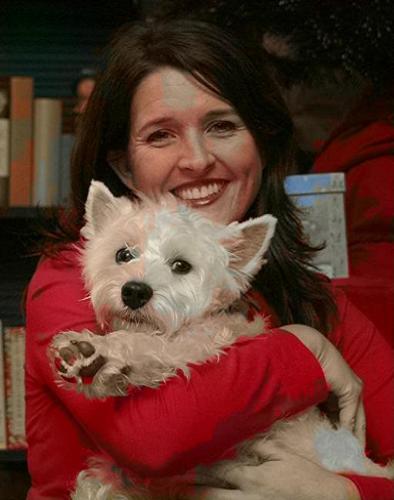}&\includegraphics[width=3.0cm]{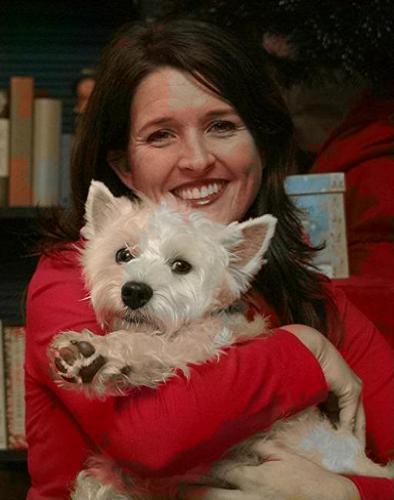}&\includegraphics[width=3.0cm]{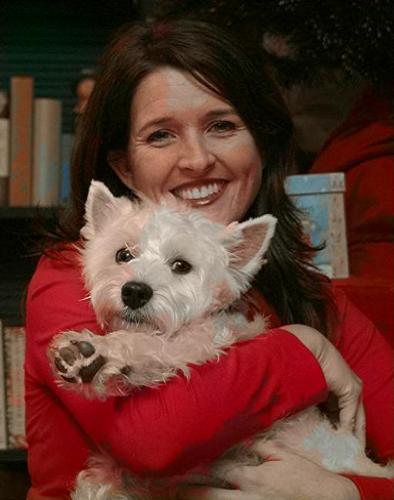}\\ 
\hline
& $1,170$ bytes & $195$ bytes & $279$ bytes & $427$ bytes\\
\includegraphics[width=3.0cm]{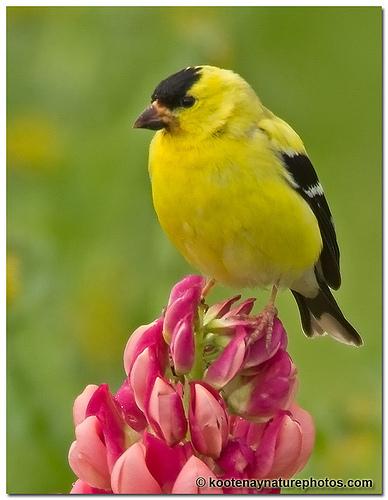}&\includegraphics[width=3.0cm]{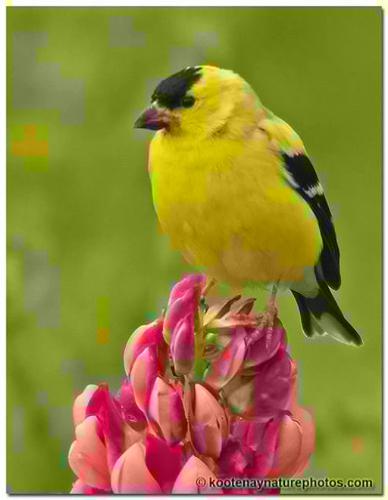}&\includegraphics[width=3.0cm]{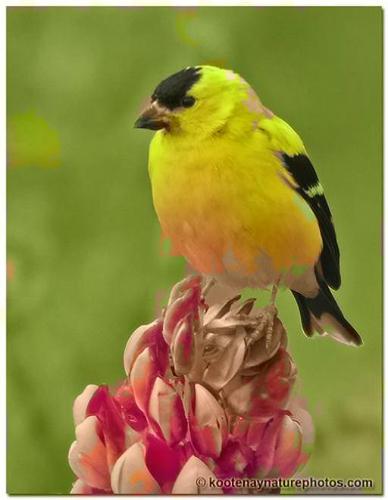}&\includegraphics[width=3.0cm]{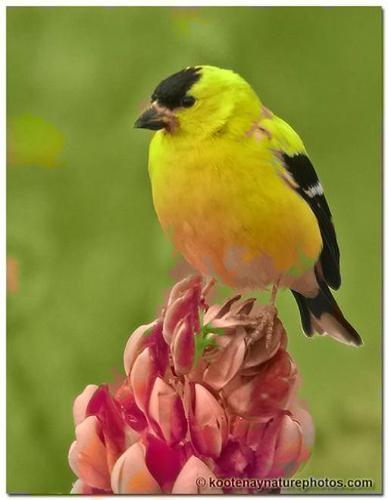}&\includegraphics[width=3.0cm]{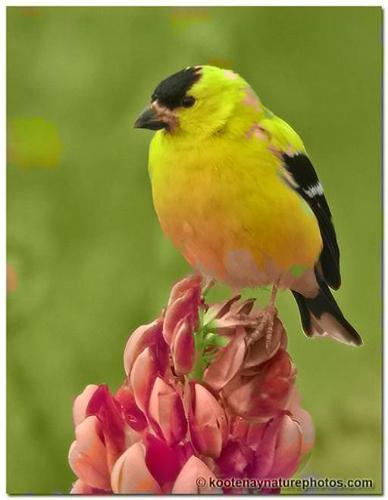}\\ 
\end{tabular}
\caption{We show visualizations generated by our proposed low-cost framework and JPEG color coding~\cite{jpeg} (for JPEG we report the storage space required to compress {\em only} the color channels). Our approach, produces vibrant realistic looking images at only about 1/$6^{th}$ of the storage size required by JPEG.}
\end{table*}

Learning to colorize grayscale images is an important task for three main reasons. First, in order to predict the appropriate chroma of objects in an image, a colorization model effectively learns to perform high level understanding from {\em unlabeled} color images. In other words, it learns to recognize the spatial extents and the prototypical colors of semantic segments in the picture. Since unlabeled photos are plentiful, colorization can be used as an {\em unsupervised} pre-training mechanism for subsequent supervised learning of high-level models for which labeled data may be scarce. Second, colorization can be useful for artistic pursuits by giving new life to grayscale vintage photos and old footage. Finally, colorization models can greatly help with image and video compression. Most objects cannot have all possible colors and by learning the plausible color space for each object we can more compactly encode the color information. 
In this work we focus predominantly on this last application of colorization, by learning parametric models of image colorization for image compression. We outline the challenges posed by colorization for image compression and propose a new deep architecture to overcome these hurdles.	

Recent successful learning-based approaches~\cite{zhang2016colorful,larsonLearningRepresentations} for automatic colorization operate under the regime of ``zero-cost,'' i.e., they assume that the output color must be predicted from the input grayscale image without any additional storage expense. While this may be reasonable for generating artistic colorization automatically, it is not applicable for the purpose of image compression as many objects in the real world admit multiple plausible colors. The problem is exemplified in Table~\ref{tb:comparingApproaches} where we report zero-cost colorization results for different methods as well as our approach. While some of the colors produced by these methods are realistic-looking, they are actually quite different from the ground truth (first column). In order to reproduce with high fidelity the true colors, we propose to 
store some additional information that helps to disambiguate between the choices (last column of Table~\ref{tb:comparingApproaches}).

To account for the multimodal color distribution of many objects we propose a convolutional neural network (CNN) that takes as input a grayscale photo and outputs $K$ plausible color values per image pixel, where $K$ is treated as a hyper-parameter defining the complexity of the model. The multiple outputs are produced by using a CNN structured in the form of a tree, with a single trunk splitting at a given depth into $K$ branches, each generating a candidate color per pixel. The trunk contains convolutional layers that compute shared features utilized by all branches, while each individual branch predicts a distinct plausible color mode for each pixel. We study how to use this architecture both in the zero-cost setting as well as for compression. In the zero-cost setting, we train the network to choose one of its $K$ \difftxt{candidate} outputs at every pixel. We then discuss how the multimodal output of our CNN can be leveraged to perform highly effective image color compression. As can be observed in Table~\ref{tb:teaser}, our approach can generate colors that are virtually indistinguishable from the ground truth at the storage cost of merely a few hundred bytes for a high-resolution picture. To achieve the same feat the JPEG~\cite{jpeg} codec requires an order of magnitude more bytes.

In summary, these are the contributions of our work:
\begin{enumerate}[(1)]
%\begin{enumerate}
\item We introduce a tree-structured network architecture that can produce multiple plausible colorizations of an input grayscale picture. 
\item We discuss how we can apply this model for zero-cost colorization, where the objective is to produce a single color hypothesis. We show that our approach is competitive with existing algorithms in this field.
\item We describe how we can leverage the multimodal output of our CNN to perform highly effective image color compression, which corresponds to the case of non-zero-cost colorization. In this regime, we study approaches that can be used to generate varying trade-offs between color fidelity and image size. 
\end{enumerate}

\section{Background}
	In order to understand the position of our work %and colorization for image compression 
	we begin with a discussion of the literature in the \difftxt{field} in two parts. First we review existing \difftxt{``zero-cost"} colorization approaches and discuss how our algorithms tackles this task differently from these prior methods. Then, we compare our approach to existing colorization approaches for compression and provide an overview of the benefits of our approach.
	
	A recent trend in zero-cost colorization approaches entails the use of deep learning to train models from images annotated with class labels indicating the objects present in the photos, such as traditional datasets for image categorization. Examples of this genre are the work of Dahl~\cite{dahl2015}, Zhang et al.~\cite{zhang2016colorful} \difftxt{Larsson} et al.~\cite{larsonLearningRepresentations} and Iizuka et al.~\cite{IizukaSIGGRAPH2016}. These approaches learn features from grayscale images that are useful for producing colors. In contrast, Cheng et al.~\cite{deepColorization} learn a network for colorization on top of existing feature descriptors.
	Only some of these methods~\cite{zhang2016colorful,larsonLearningRepresentations} handle multi-modality of image colors explicitly by predicting probability distributions over a quantized output space. 
	
These modern parametric models are in contrast to ``color transfer'' methods such as the work by \difftxt{Charpiat et al}.~\cite{charpiat2008automatic} and  \difftxt{Deshpande et al.~\cite{desphande2015}}. Color transfer methods work by finding color images similar in content to a grayscale testing input in a large database and by transferring color from these reference images using local features. These approaches address the issue of multi-modality of color output by relying on finding example images in large databases with color distributions similar to the ground-truth color image. 
Some other approaches use human annotations to identify color seeds or image regions. Chia et al.~\cite{chia2011semantic} use hand-labeled image regions from \difftxt{grayscale} images to search the internet for similar semantic content and to identify plausible colorizations. Levin et al.~\cite{levin2004}  proposed the use of human labelling of color seeds which are propagated to non-seed pixels for dense color output.

Our approach differs from existing \difftxt{``zero-cost"} colorization approaches in how we model the multi-modality of object colors in the real world. We use a tree-structured network that branches at a certain layer to predict multiple color hypotheses for each pixel. Our model is trained to reproduce exact colors rather than ``color classes", as instead done in \difftxt{Larsson} et al.~\cite{larsonLearningRepresentations} and Zhang et al.~\cite{zhang2016colorful}. To produce a final zero-cost colorization we train a separate module to identify the most plausible color hypothesis for each pixel as a secondary step. 
	
	\difftxt{Whereas there have been recent efforts in image compression with deep networks such as the work of Toderici et al.~\cite{toderici2015variable}, these efforts focus on generating the entire image as opposed to colorizing a grayscale input. Furthermore, Toderici et al.~\cite{toderici2015variable} operate on images at very low resolutions ($32 \times 32$)  since generating complete images at large resolutions is very hard. Instead, our approach operates on full resolution images.}	
	
	We now turn to a discussion of  prior approaches to the problem of colorization for compression. Some of the methods in this genre~\cite{cheng2007learning, he2009unified} operate by storing the true colors of a small subset of pixels (the ``seeds'') and then apply color propagation techniques~\cite{levin2004} to extend color information to all the other pixels in the image. 
	
	Egge et al.~\cite{egge2015cfl} propose a low-cost colorization method based on fitting a parametric regression model to each grayscale patch of the image to approximate its colors. The parameters \difftxt{of the model for each patch} are stored and used at decoding time to generate the colors. 
		
	Our work differs from these prior approaches as it approximates the true colors by performing an analysis of the semantic content of the image, rather than by fitting a parametric model to each patch or by storing a subset of colors.
	
	\difftxt{Recent deep learning based approaches~\cite{zhang2016colorful,larsonLearningRepresentations,IizukaSIGGRAPH2016,dahl2015} contribute to the field of colorization by showing that it is possible to produce good colors by analysis of the content of the images. One key problem faced by analysis based synthesis approaches is handling the multi-modality of colors in objects. One approach adopted by Larsson et al.~\cite{larsonLearningRepresentations} and Zhang et al.~\cite{zhang2016colorful}	involves posing the colorization problem as a classification task. This prevents the color averaging problem faced by models trained for regression. In contrast, we handle multi-modality in the output space with a novel tree structured architecture. We show that this architectural modification for handling multi-modality helps improve performance on compression and zero-cost colorization. The tree structure allows us to handle multi-modality while training for regression (instead of classification). Classification based approaches are limited in their ability to accurately reproduce the colors (because they assign one of the quantized color classes to each pixel) whereas with regression we can in principle predict accurately the color value. In the zero-cost colorization setting our model also produces a single color output with the addition of a module that chooses which branch to use for which pixel, however as opposed to classification based approaches our color classes vary from pixel to pixel making our model substantially more expressive.}
	The ability to generate accurate color outputs is crucial in the compression scenario, where the objective is to reproduce faithfully the original colors with minimal additional information. In this low-cost setting, we demonstrate substantial improvements over \difftxt{JPEG}~\cite{jpeg}, which represents the most commonly used image compression codec. 

\begin{figure*}[!ht]
\centering
\includegraphics[width=14.0cm]{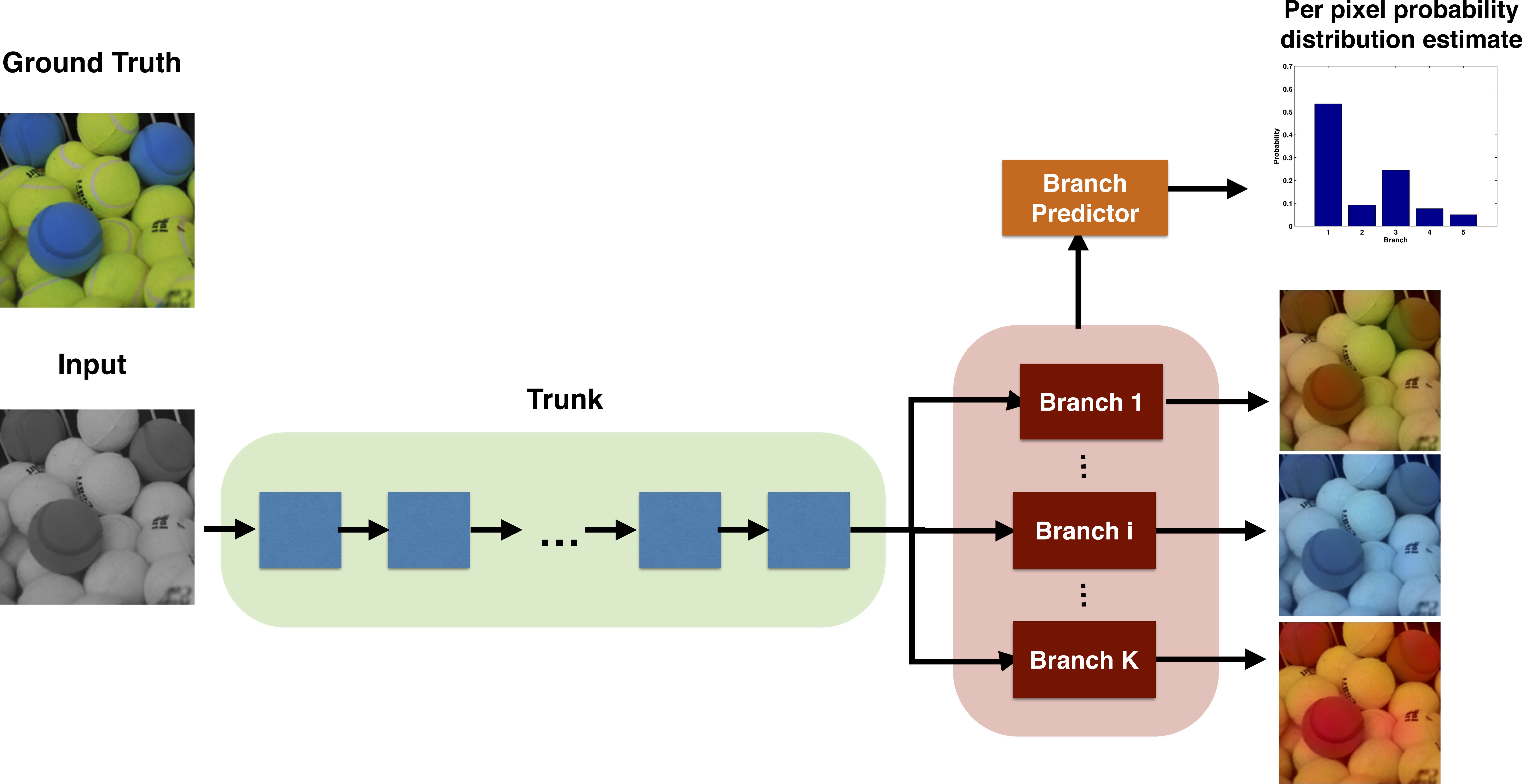} \\
    \caption{High-level illustration of our tree-structured network architecture. Each branch outputs one color value per pixel. This allows our model to produce $K$ hypotheses for each pixel. The branch predictor estimates the best performing branch for every pixel. This can be used for zero-cost colorization, where a single color output (the most likely) must be predicted.}
\label{fig:branching}
\end{figure*}
		
\section{Technical Approach}
	
	We adopt a two-step approach for image compression via colorization. First we use our proposed branched deep network to generate multiple color hypotheses for each pixel. Second, we specify how to compactly store the best hypothesis at a per-pixel level so as to produce low-cost colorization at inference time. 

	We start by formally defining our colorization problem. Colorization entails generating a color image from a grayscale image. We assume we are given a training set of $N$ examples, where $I_G^{(i)} \in \mathbb{R}^{H \times W}$ represents the grayscale version of the $i$-th training image and $I_C^{(i)} \in \mathbb{R}^{H \times W\times 2}$ contains its two color channels. Our task is to estimate $I_C^{(i)}$ from $I_G^{(i)}$. 
	
	 $I_C$ and $I_G$ can be represented in many different ways. Cheng et al.~\cite{deepColorization} make use of the YUV color space. \difftxt{Deshpande et al.~\cite{desphande2015}}, Zhang et al.~\cite{zhang2016colorful} and Iizuka et al.~\cite{IizukaSIGGRAPH2016} adopted the LAB color space, \difftxt{whereas Larsson et al.~\cite{larsonLearningRepresentations} adopt the Hue/Chroma space}. Since our work is geared towards image compression, we make use of the $YC_bC_r$ color space, which is more commonly used by image compression architectures.

	We begin by considering a potential learning objective for our colorization task. The goal is to learn a model \difftxt{${\mathcal F} (\cdot ; \theta)$} \difftxt{parametrized} by weights $\theta$ that can be used to predict the color channels associated with the grayscale input, i.e., such that ${\mathcal F} (I_G; \theta) \approx I_C$. A seemingly natural choice for the learning objective is the sum of squared L2 distances between predicted color values and the ground truth, i.e., 

\begin{equation}\label{eq:colorizationObjective}
	E(\theta) = \sum\limits_{i=1}^{N} \sum\limits_{(x,y)} || {\mathcal F} (I_G^{(i)};\theta)|_{(x,y)} - I_C^{(i)}|_{(x,y)} ||_2^2
\end{equation}
		
This objective guides the model to learn that only $I_C^{(i)}$ is the correct color estimate for a given input $I_G^{(i)}$. However, we know that this is not true based on our observation of the real world where many objects have multimodal color distributions. Our experiments suggest that a model trained with this objective learns to reproduce fairly well the colors of objects that have unique or characteristic chroma, such as the sky. But this model is reluctant to predict a vibrant color hypothesis for objects that do not have a single prototypical chroma. In such cases the model avoids committing to a clear response and instead tends to predict a conservative ``average color'' that has small L2 distance from most possible values. This phenomenon is clearly visible in the comparative analysis between different approaches shown in Table~\ref{tb:comparingApproaches}. The second column shows the results achieved with the system of Dahl~\cite{dahl2015}, which is trained to minimize Eq.~\ref{eq:colorizationObjective}. This leads to \difftxt{``grayish"} estimates on highly multi-modal color regions such as clothing but predicts correct colors for unambiguous regions, as can be seen in the nearly exact color reproduction for sky and grass.

\subsection{\bf Learning Multiple Color Hypotheses with Branching}	 

	In order to address the inherent ambiguity of colorization, we propose a network architecture that predicts $K$ color hypotheses per pixel. This is achieved by means of a CNN whose single trunk splits at a certain depth into $K$ distinct branches, each outputting a 2-channel color output per pixel. All layers, both in the trunk as well as in the $K$ branches, are fully convolutional. The rationale behind this architecture choice is that it allows us to express the multi-modal plausible colors of many objects in the real world with a reduced number of parameters. 	Rather than training $K$ distinct networks, which would require a huge number of parameters and would have large computational cost, a single tree-structure network with a shared trunk (capturing common features) is parsimonious both in terms of storage as well in terms of training cost. Furthermore, a single optimization (as opposed to disjoint training of separate networks) assures that the $K$ branches will naturally diversify in order to cover the multi-modality of the output. Figure~\ref{fig:branching} illustrates the proposed branched architecture.
	
\begin{table*}[t!]
\centering
\setlength{\tabcolsep}{1pt}
\begin{tabular}{c|ccccc}
\toprule
 \difftxt{Original} & Branch 1 & Branch 2 & Branch 3 & Branch 4 & Branch 5 \\ 
\hline
\includegraphics[width=2.9cm]{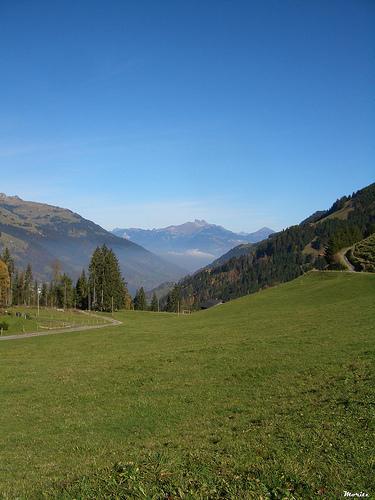}&\includegraphics[width=2.9cm]{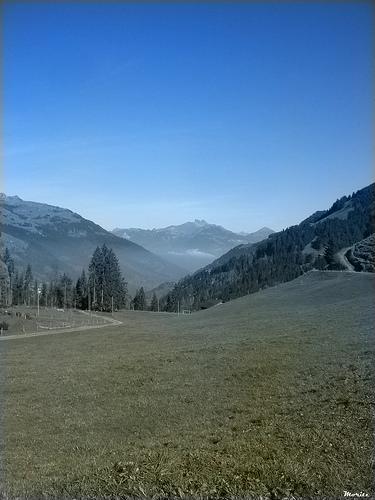}&\includegraphics[width=2.9cm]{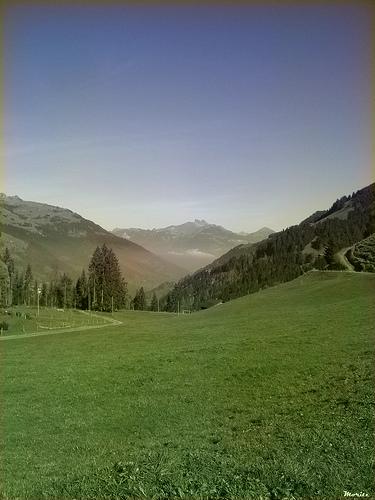}&\includegraphics[width=2.9cm]{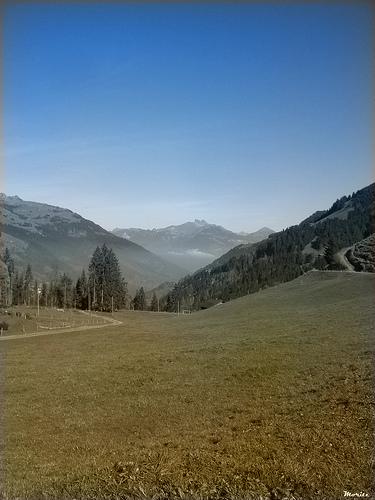}&\includegraphics[width=2.9cm]{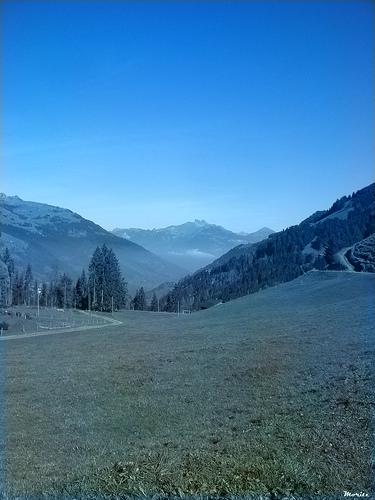}&\includegraphics[width=2.9cm]{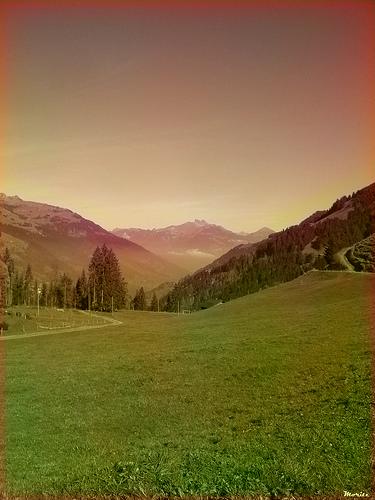}\\ 
\includegraphics[width=2.9cm]{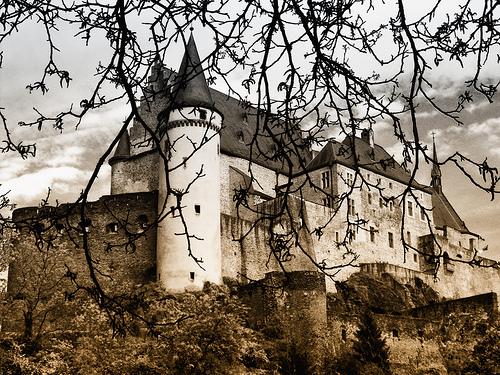}&\includegraphics[width=2.9cm]{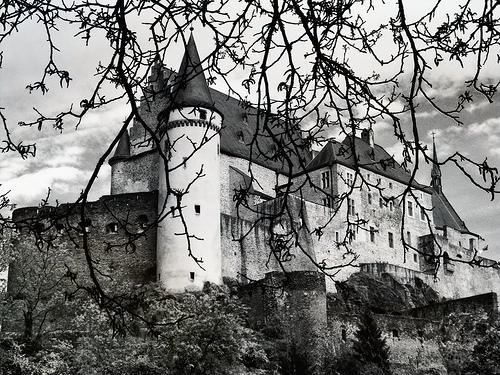}&\includegraphics[width=2.9cm]{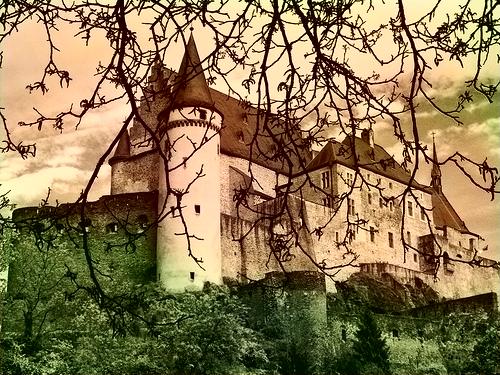}&\includegraphics[width=2.9cm]{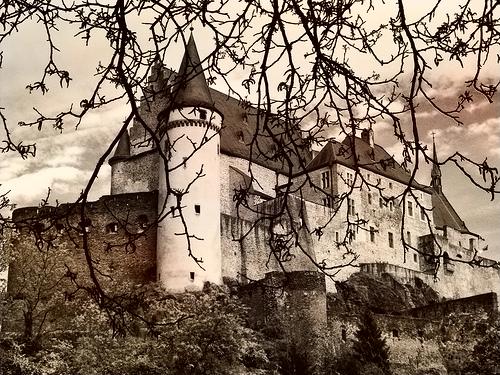}&\includegraphics[width=2.9cm]{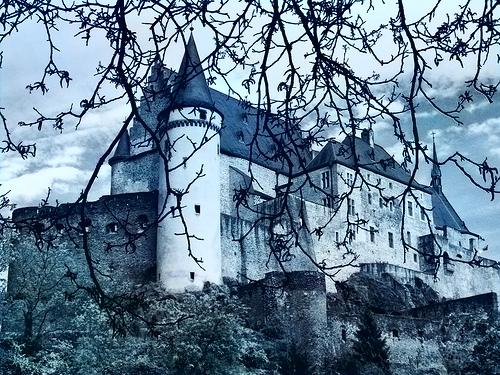}&\includegraphics[width=2.9cm]{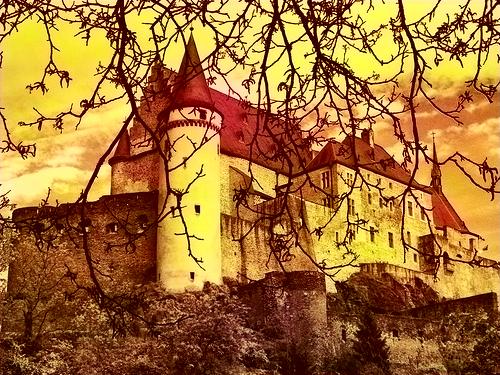}\\ 
\includegraphics[width=2.9cm]{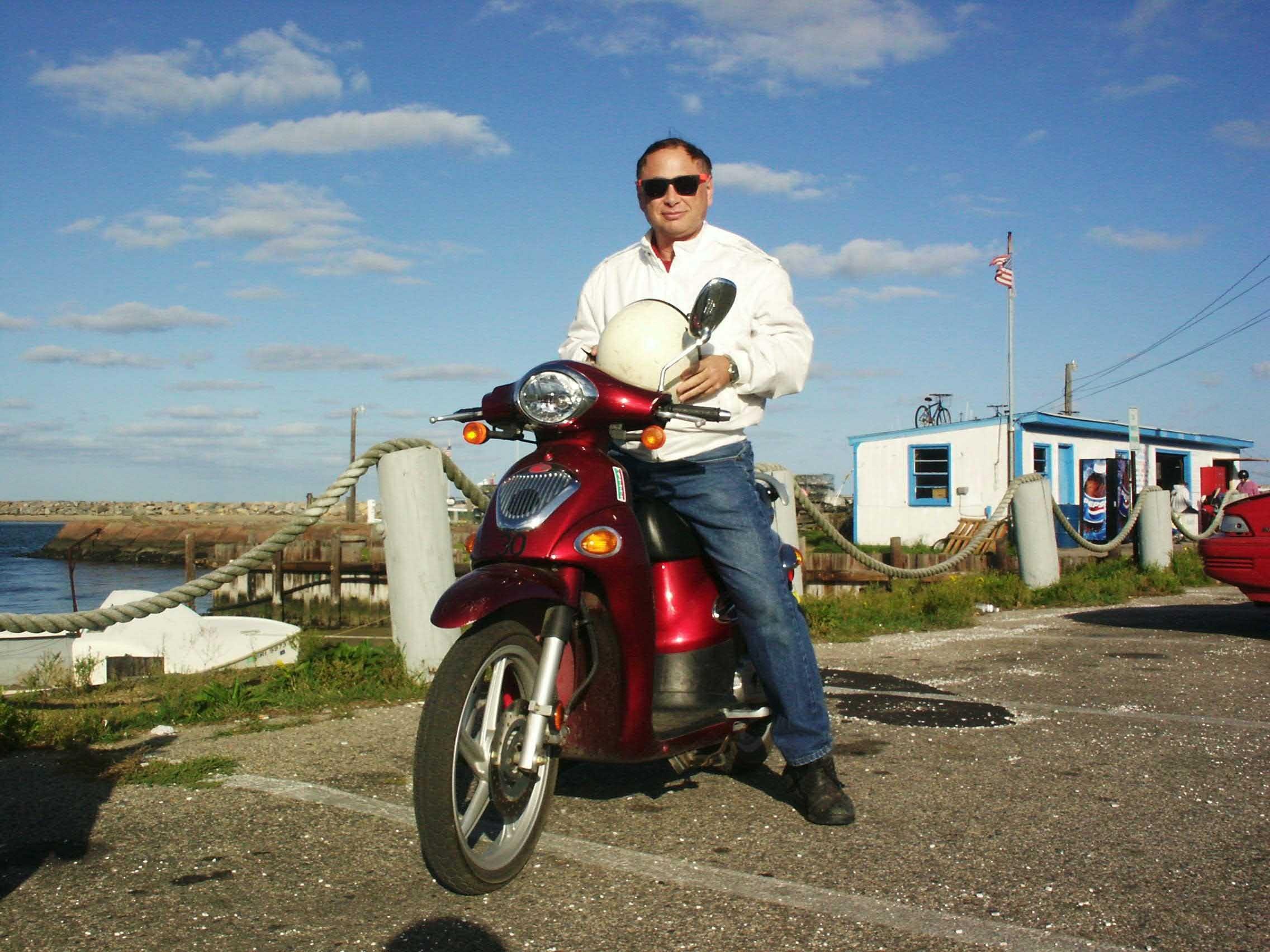}&\includegraphics[width=2.9cm]{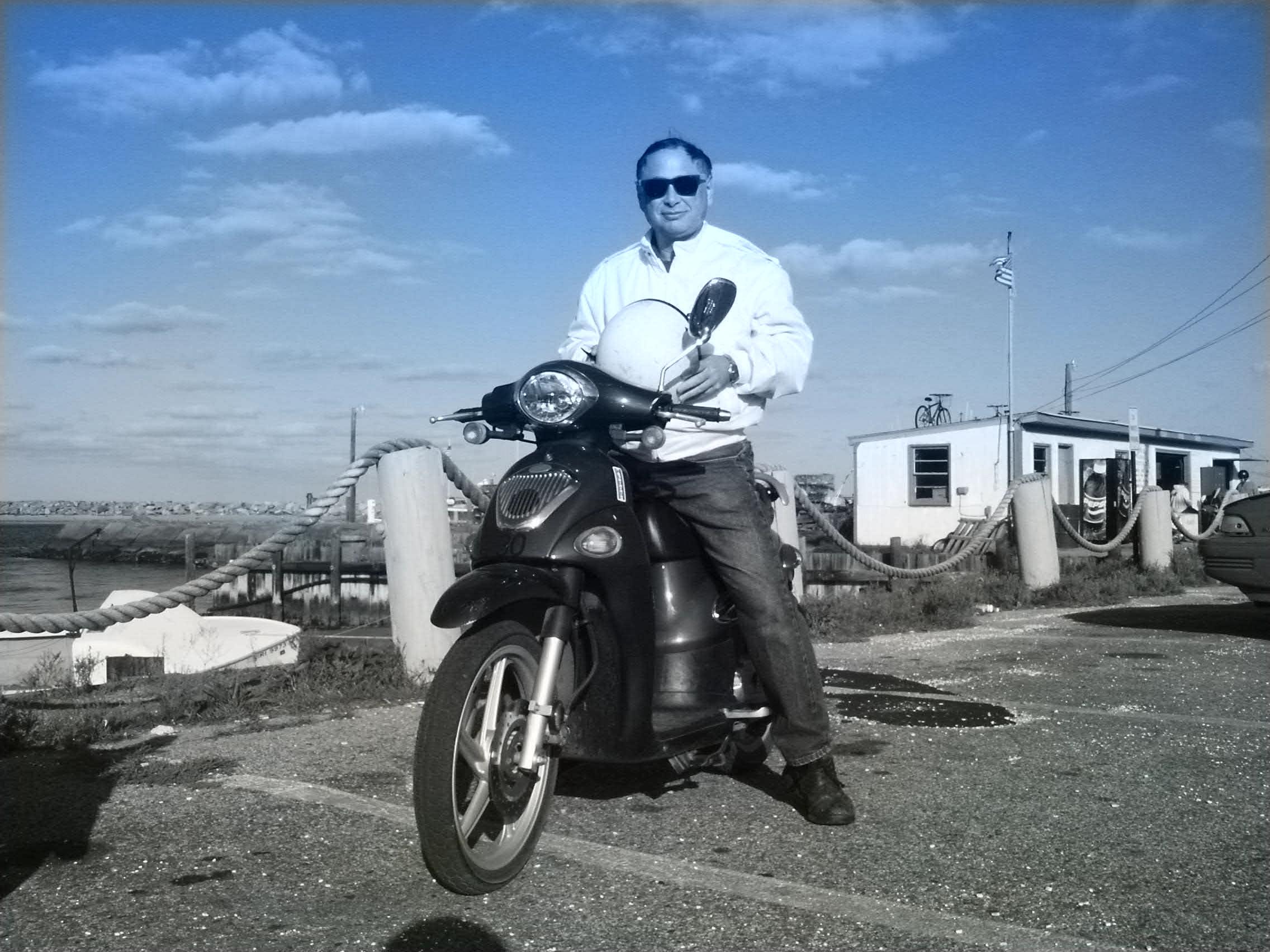}&\includegraphics[width=2.9cm]{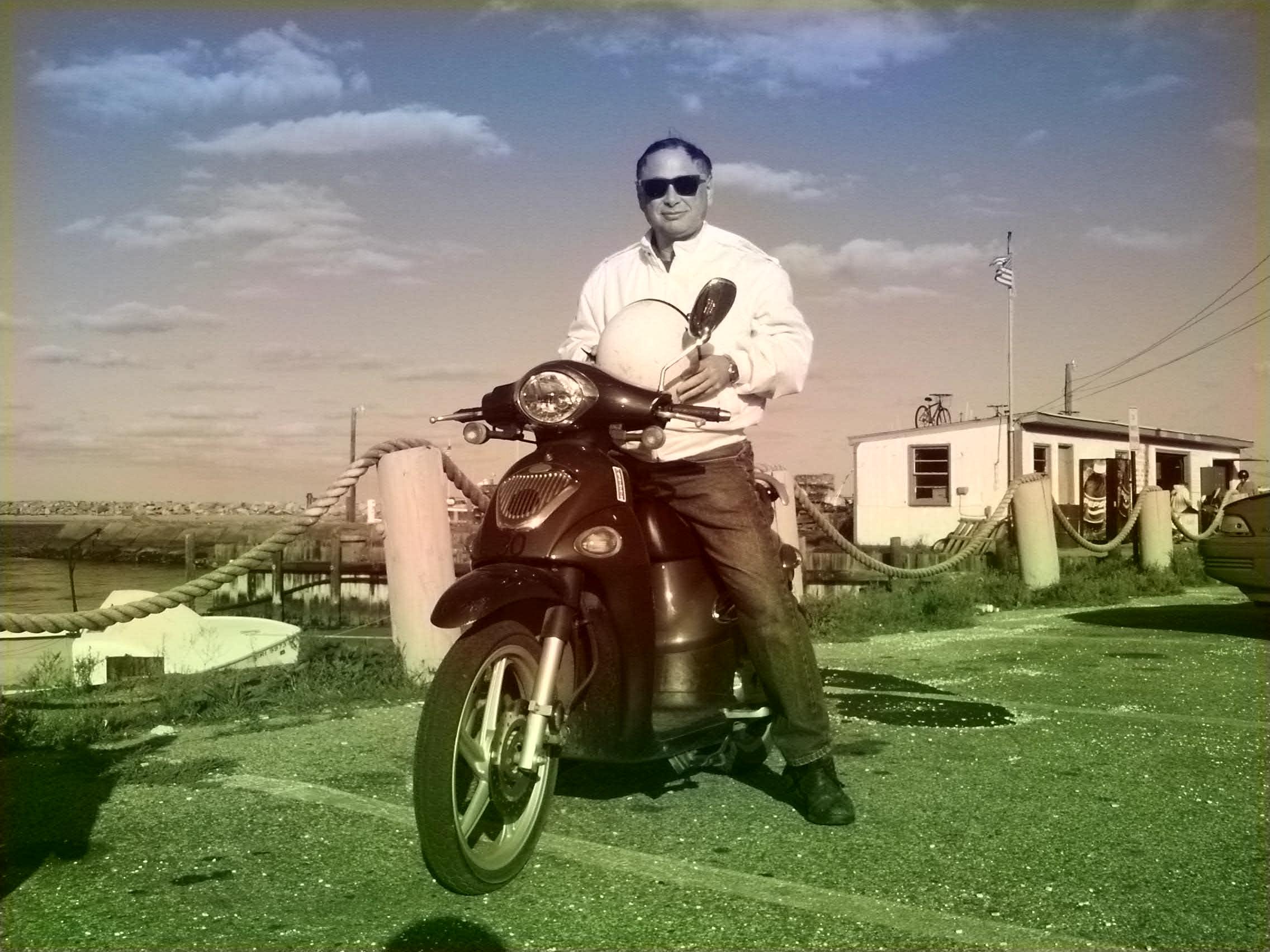}&\includegraphics[width=2.9cm]{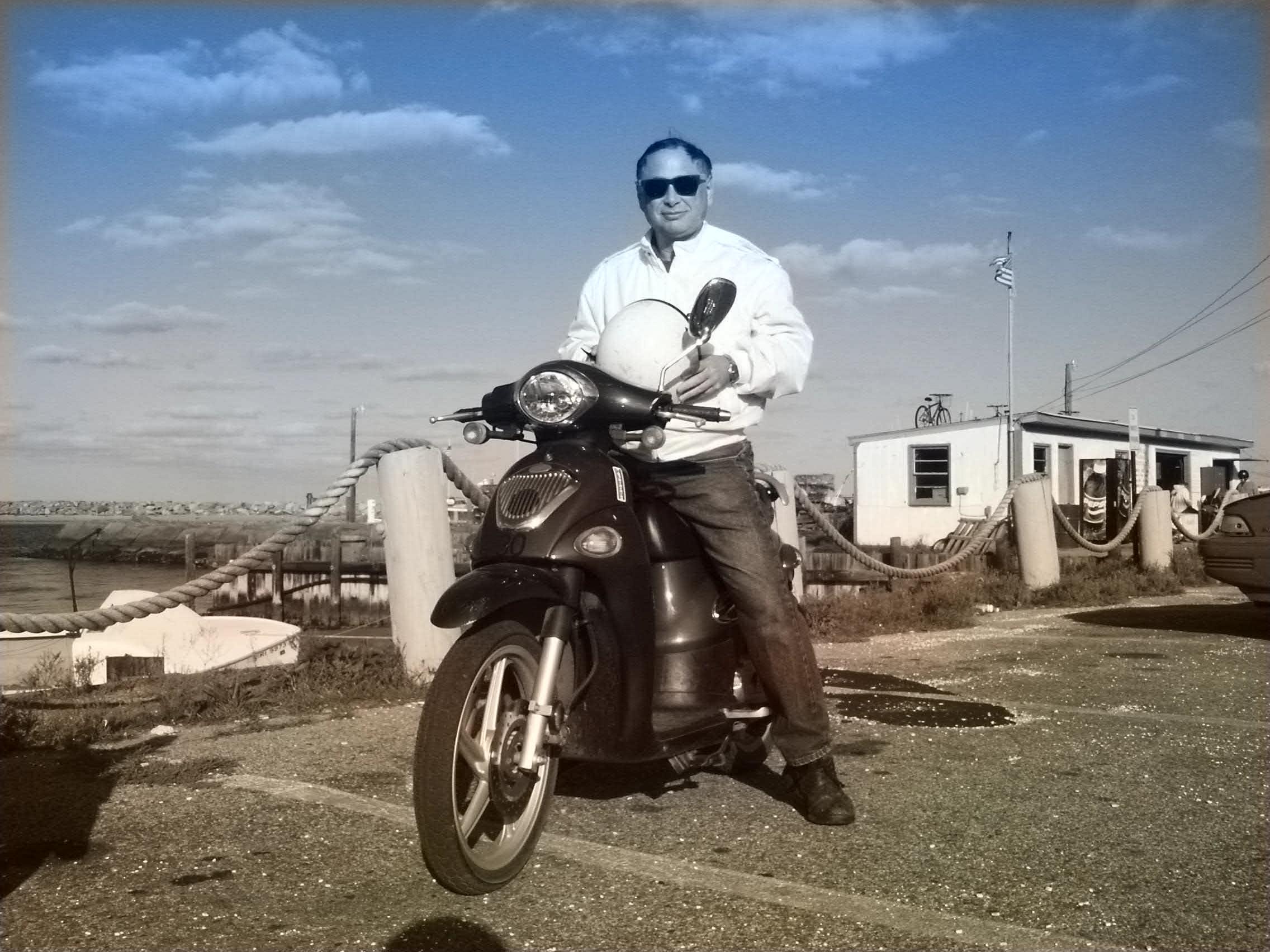}&\includegraphics[width=2.9cm]{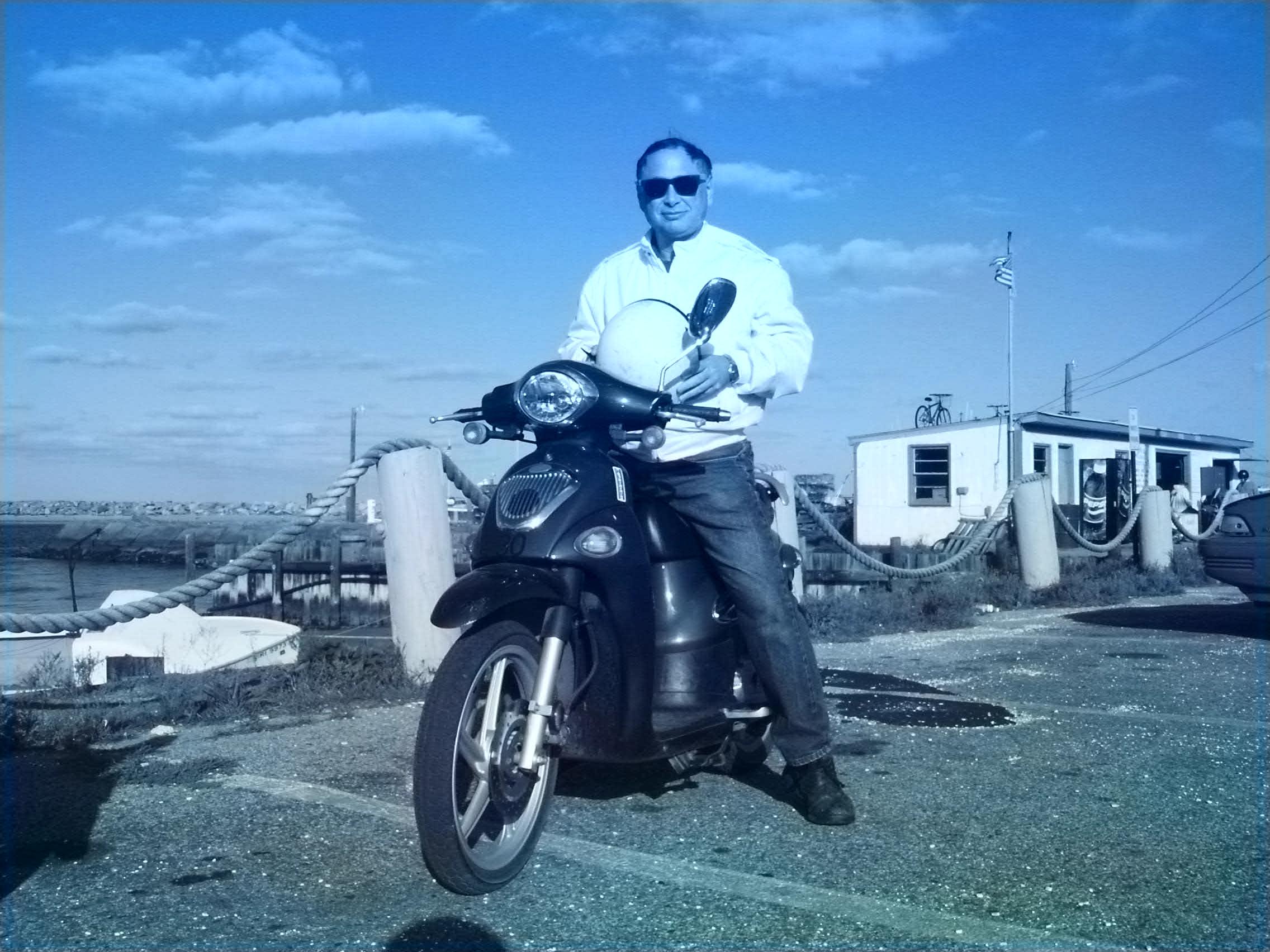}&\includegraphics[width=2.9cm]{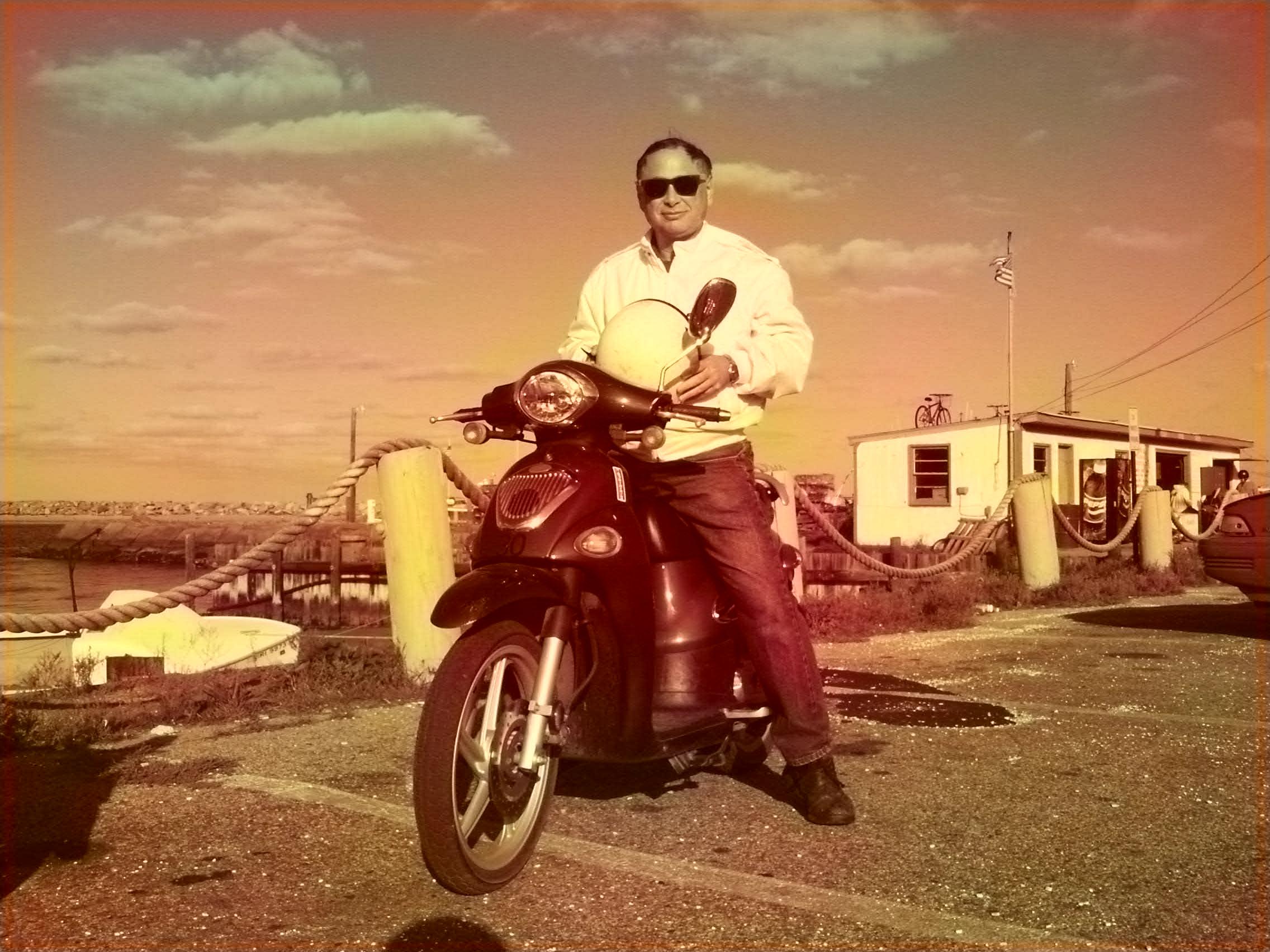}\\  
\hline
\end{tabular}
\caption{We show the output produced from each of the $K=5$ branches when presented with a grayscale input. The branches estimate various plausible shades of colors for objects such as grass and sky within the image. These shades reflect color variations observed for these objects in the training set. \difftxt{More specifically we can see that Branch 3 tends to colorize grass and bushes as brown (a plausible color) whereas Branch 2 colorizes them as green. Both branches are implicitly able to recognize and localize grass and plants and propose appropriate colors for them without ever being explicitly taught the notion of grass.} }\label{tb:diverseoutputs}
\end{table*}

	The hyperparameter $K$ controls the complexity of our model. Its  value depends on the number of color modes exhibited by objects in the application domain. The greater the value of $K$, the higher the number of color hypotheses that the model can express. But a higher value of $K$ also increases the number of parameters that must be learned. In the experiments we report empirical evaluations for various values of $K$.
	
	The question is: how do we train this multi-output architecture given that for each training grayscale input $I_G^{(i)}$ we are only given one color version of it, its ground-truth color channels $I_C^{(i)}$? We assume that the color of each pixel in $I_C^{(i)}$ must be generated by {\em one} of the $K$ branches. But different pixels may be generated by different branches. In other words, the color of each pixel is generated independently from the others. During each iteration of training, for each image in the mini-batch we perform forward propagation through all branches to produce $K$ color values for each pixel. Then, we assign to each pixel the branch that best approximates the color of that pixel. Finally, the backpropagation update for that pixel will only change the branch assigned to the pixel. Thus, the loss optimized by our model is of the form:
	
	\begin{equation}\label{eq:branchLoss}
	{\hat E}(\theta) = \sum\limits_{i=1}^{N} \sum\limits_{(x,y)} \min_{j=1\dots K}|| {\mathcal F}_j (I_G^{(i)};\theta)|_{(x,y)} - I_C^{(i)}|_{(x,y)} ||_2^2
\end{equation}
	
	where ${\mathcal F}_j (I_G^{(i)};\theta)$ represents the color output  produced by the $j$-th branch in our network. Note that this training scheme allows us to learn models without predetermined manual assignment of pixels to branches. Although we update a single branch for every pixel, because every mini-batch contains several images, each consisting of many pixels, in practice we typically update all branches in every mini-batch iteration. A similar scheme of training was adopted by Vondrick et al.~\cite{vondrick2015anticipating} to learn to anticipate multiple hypotheses of future actions from unlabelled video. 

\begin{table*}[ht!]
\centering
\setlength{\tabcolsep}{1pt}
\begin{tabular}{cc|cc|ccc}
\toprule
  & & \multicolumn{2}{|c|}{Grid} & \multicolumn{2}{|c|}{Segmentation} \\
 \midrule
Image & Oracle  & \multicolumn{1}{|c}{Small} & \multicolumn{1}{c|}{Large} & \multicolumn{1}{|c}{Small}  & \multicolumn{1}{c|}{Large} \\
\midrule
 & & 137 & 419 & 152 & 601\\
\includegraphics[width=2.8cm]{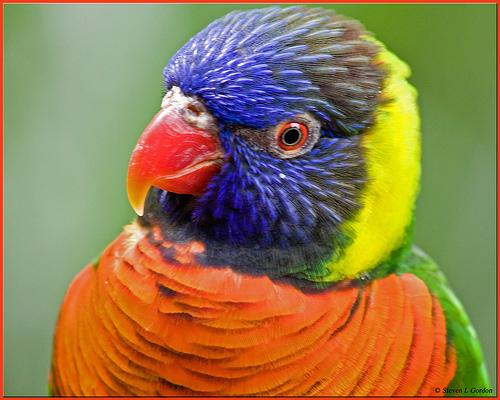} &
\includegraphics[width=2.8cm]{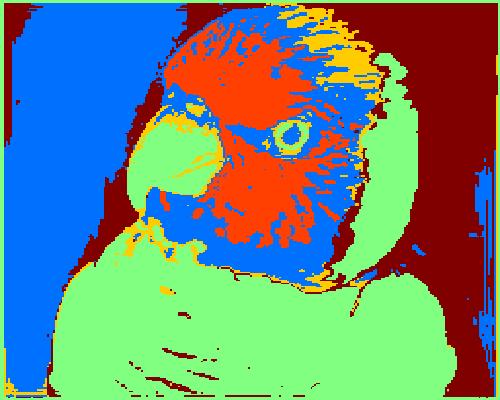} & \includegraphics[width=2.8cm]{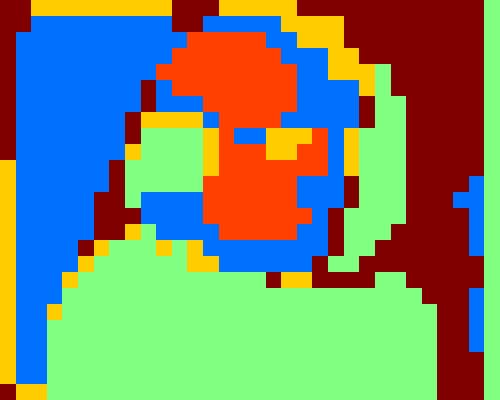}  & \includegraphics[width=2.8cm]{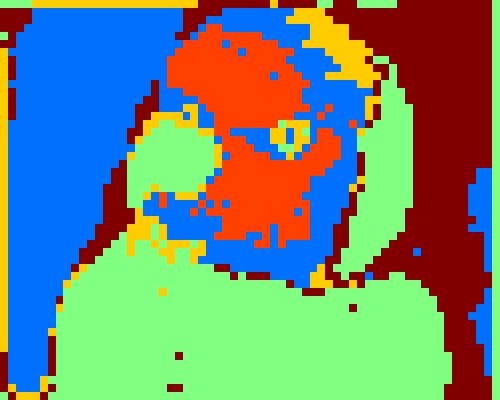} & \includegraphics[width=2.8cm]{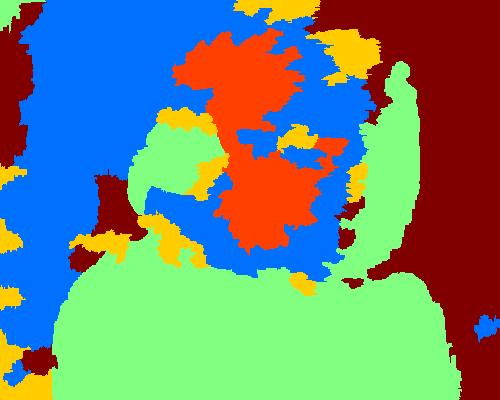}  &   \includegraphics[width=2.8cm]{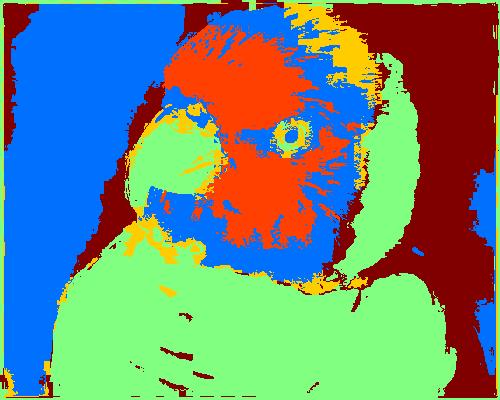} \\
 & & 110 & 894 & 168 & 418\\
\includegraphics[width=2.8cm]{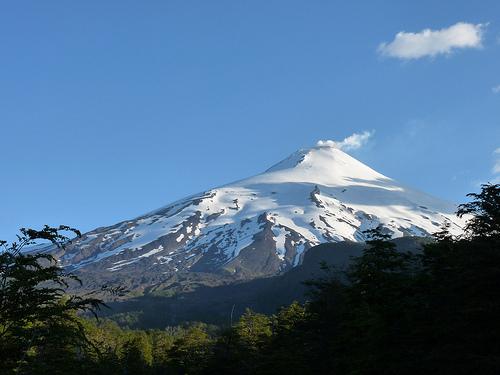} &
\includegraphics[width=2.8cm]{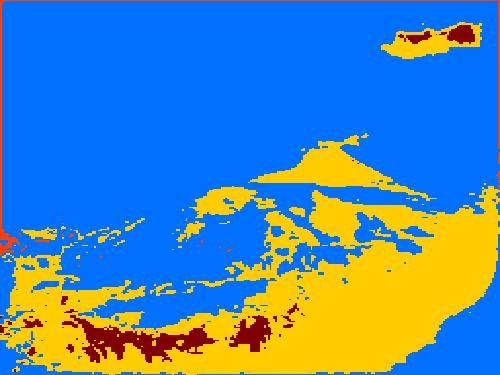} & \includegraphics[width=2.8cm]{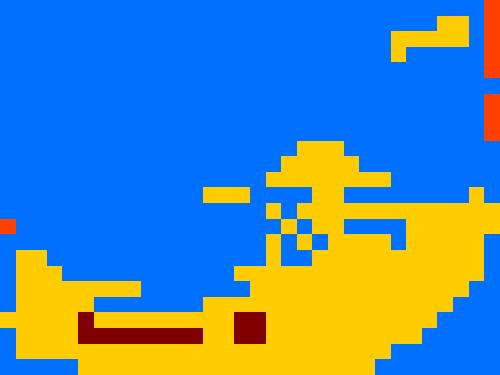}  & \includegraphics[width=2.8cm]{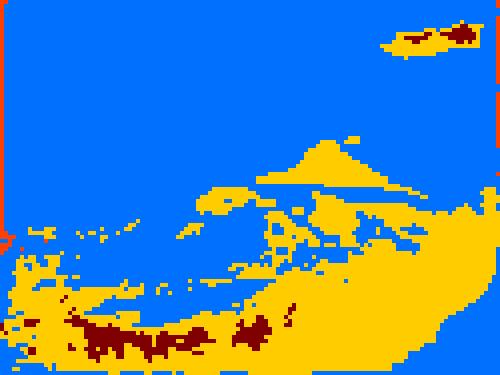} & \includegraphics[width=2.8cm]{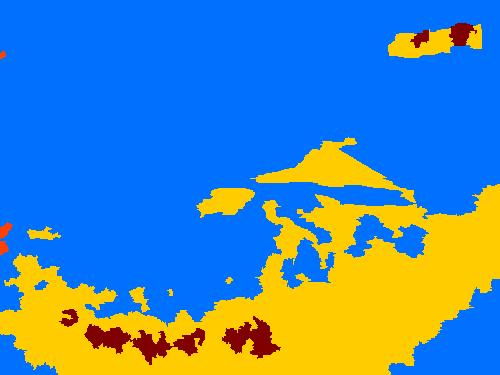}  &  \includegraphics[width=2.8cm]{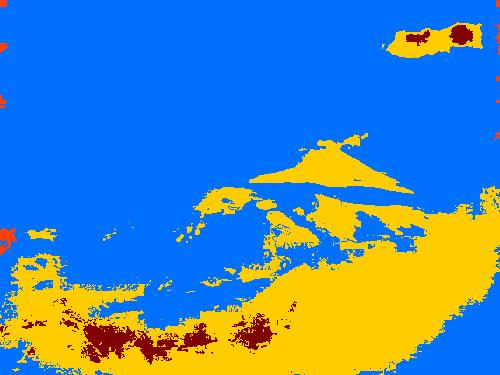} \\
\end{tabular}
\caption{The second column shows the oracle encoding in the form of a colormap of the branch indices. Columns 3-6 represent various approximations to the oracle encoding by exploiting region coherence, based on a subdivision of the image into cells (columns 3-4) or segments (columns 5-6). Above each encoding we state the size in bytes required to store the decisions with the choice of encoding. }\label{fig:spatialCoherrence}
\end{table*}

	Note that while a single-branch architecture will tend to predict bland ``average colors'' for ambiguous regions (in an attempt to keep the L2 error low), our model will naturally exploit the multiple branches to produce a diverse set of color hypotheses representing distinct color interpretations of the objects. This can be seen in  Table~\ref{tb:diverseoutputs} where we visualize the color image produced by each branch. It can be noted that the  branches generate diverse outputs representing multiple color hypotheses for each pixel. \LT{This diversification is the  result of our loss function (Eq.~\ref{eq:branchLoss}), which measures error {\em only} with respect to the optimal branch for each pixel. Because there is no penalty associated to the incorrect branches, the network is implicitly encouraged to produce $K$ disparate and ``bold'' (rather than conservative) predictions. This can be better understood by considering as a simple analogy a game that we have all played as children: guessing a number, say, between 0 and 10. If we are told that we have a single chance and that the winner is the player whose guess is closest to the secret number, then we may be tempted to choose the conservative answer of ``5'', since this number minimizes the average error over all possible values of the secret number. This is similar to the ``conservative'' average-color predicted by the single-branch network when posed with the problem of guessing the color of highly ambiguous pixels. However, if in our game of guessing the secret number, we are told that we have 3 chances, and that the error is computed only with respect to our best guess, we have an incentive to guess numbers that are ``disparate'' in order to maximize our chances, e.g., ``2'', ``5'' and ``8''. This is effectively what our multi-hypothesis network does, with the only difference that its $K$ color predictions will vary for every pixel, based on the  most plausible colors estimated from analyzing the grayscale region around the pixel. One can also think of our approach as performing compression by storing an indexed colormap where the colors are generated by an analysis of the content (as opposed to traditional quantization based approaches with a fixed color palette).}

\subsection{\bf Image Compression via Low-Cost Colorization}

\paragraph{\difftxt{Oracle}}
We now consider how to leverage our colorization model to perform color compression. Note that in the compression regime we are allowed to store some additional information in order \difftxt{to} reproduce the original color of the image. A simple solution is to store an index to the branch that best approximates the ground truth color  for every pixel. This allows us to encode (approximately) the color of an image by storing $\mathcal{O}(\log_2 K)$ bits per pixel in addition to the grayscale version of the image $I_G$. Then at decoding time, we pump the grayscale version of the image through the network to produce the $K$ color outputs and use the stored branch indices to select the appropriate branch for every pixel. In the context of compression, we see that the value of $K$ influences the storage required per pixel. With fewer branches we get improved compression as fewer bits are needed to store the branch index. But with a greater number of branches, more hypotheses are generated and thus  a more accurate color approximation is expected. 

We refer to the procedure of selecting the best branch per-pixel by looking at the ground truth as using an ``oracle''. We note that the oracle encodes for each pixel the branch that best approximates the channels ($C_b$ and $C_r$) jointly. This is a design choice we made supported by experiments where we found that this strategy produces nearly the same error as when storing the best branch for each channel separately, but it allows us to cut by half the amount of the information to store, thus producing a huge saving in file size.

\begin{figure}[h]
  \centering
  \includegraphics[width=8.0cm]{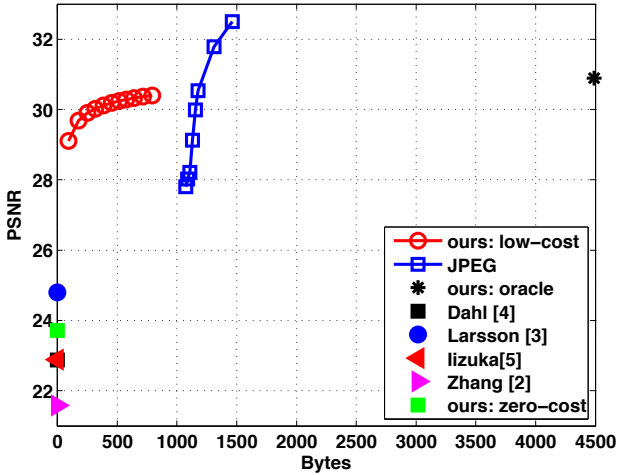}
  \caption{Compression via colorization. The plot shows number of bytes needed to store color given the grayscale image versus PSNR (quality of the final image, the higher the better). Our approach stores the branch map produced by the oracle either in uncompressed from (oracle) or in a lossy fashion using region subdivision. This produces dramatic compression savings over the standard JPEG codec for the same level of color quality for very high compression settings. We also include results of zero-cost colorization (0 bytes) with our model and other recently proposed colorization methods.}
  \label{fig:compression}
\end{figure}

As can be noted from the results in Figure~\ref{fig:compression} our oracle can \difftxt{approximate} the true colors with low error, at the expense of a small additional storage. 

\paragraph{\difftxt{Compact Lossy Encoding of the Oracle}}
\label{sec:lossOracle}
Table~\ref{fig:spatialCoherrence} shows in the second column the oracle encoding of the best branch for a few image examples. It can be observed that the oracle branch-index maps exhibit high spatial coherence. This happens because the best branch for nearby pixels is often the same branch. This suggests that in order to save storage, rather than saving the best branch index for every pixel, we can store the best branch index per-region or per-patch. Specifically, given a region, we store the index of the branch that yields the best color reconstruction with respect to the ground truth in that entire region (i.e., the branch yielding the lowest mean squared error (MSE) in the region). We consider two ways of defining regions for compression: (1) a subdivision of the image into a grid of fixed-size square cells and (2) a segmentation of the grayscale photo into superpixels using traditional bottom-up segmentation methods~\cite{quickshift,slic}. Obviously, the higher the number of regions, the more accurate the encoding of the oracle will be. 

	Columns 3-6 of Table~\ref{fig:spatialCoherrence} show the effectiveness of these approaches in approximating the oracle encoding for different storage sizes. Details about the grid subdivision and the segmentation methods used to partition the image into regions are provided in the experimental section.

\paragraph{\difftxt{Global Correction}}

We found experimentally that the color maps produced by our models can be  improved by applying a global correction with respect to the ground truth color. The global correction corrects for a similarity transform (scale and translation) between the ground truth color and the estimate. In a compression scenario, we can simply store these 4 global parameters (2 for each color channel) and use them to produce more accurate color reconstructions at decoding time. \difftxt{The global correction adds very little cost in terms of storage while it gives our model the ability to account for errors due to a global shift and scaling of the colors.}

\subsection{\bf Zero-Cost Colorization}

	Although we designed our model to operate in the scenario of image compression under the regime of low-cost colorization, we now show how to use it to perform zero-cost colorization, i.e., to produce a single colorization hypothesis but without storing additional information. In principle this could be achieved by training a single-branch model. However, as already noted, such models tend to produce conservative ``average color" predictions for objects whose chroma is ambiguous. Conversely, our multi-branch architecture can produce more vibrant colors as, during training, the branches are never penalized for guessing incorrect colors, as long as for each pixel at least one of them predicts a good approximation. This intuition is confirmed by results shown in our experimental section (see, e.g., Table~\ref{tb:branching_performance}). However, in the zero-cost setting we are asked to produce a single  plausible color version of the input grayscale image $I_G$ from the multiple hypotheses generated by the branches. A simple solution would entail choosing arbitrarily one of the $K$ color outputs ${\mathcal F}_j (I_G;\theta)$ for \difftxt{$j=1,\hdots,K$}. A better strategy instead is to attempt to predict the best branch to use for each individual pixel, given that the training of the colorization model was also done by choosing the best branch for each pixel independently. This is effectively a multi-class classification problem, where for each pixel $(x,y)$ of $I_G$ we are asked to guess the best branch label $b(x,y) \in \{1, \hdots, K\}$. 
	
	To perform this branch prediction we add a separate component to the network outlined above. This component is illustrated in Figure~\ref{fig:branching}. The prediction component takes as input the feature maps from the $K$ branches and it is trained to predict a 1-hot vector encoding of the oracle. We train this additional component with the multinomial logistic loss per pixel \LT{after the learning of the colorization network.}
	
\subsection{\bf 	Features of our approach}
\difftxt{Before discussing the performance of our proposed architecture, we summarize some of the salient features of our proposed method:
\begin{itemize}
\item We introduce a new multiple-hypotheses architecture for colorization that is specifically designed to colorize images when the task is image compression.
\item We define a new learning objective that encourages diversification of the multiple outputs.
\item We show how to leverage the output of our colorization net to perform compression. We also shed light on what information is useful to store to perform image compression.
\item Our model has the ability to generate multiple plausible colorizations for a given image.
\item Despite having an architecture designed specifically for compression, we also show how to leverage our architecture to produce colorizations in the ``zero-cost" setting, where the goal is to produce a single vibrant colorization hypothesis without storing additional information. 
\end{itemize}}
	
\section{Experiments}

	We evaluate our approaches for colorization of images on two datasets, CIFAR100~\cite{cifar100} and \difftxt{ImageNet}~\cite{ILSVRC15}. CIFAR100 is small enough ($50,000$ training and $10,000$ testing examples) that it allows us to test various incarnations of our models but at the same time it is diverse enough (containing examples from $100$ classes) for the conclusions drawn to be meaningful. The downside of using CIFAR100 is that images are quite small ($32\times 32$) and span fewer classes. Thus it is difficult to tell how the results scale with respect to higher-resolution images and more varied content. We use the \difftxt{ImageNet}~\cite{ILSVRC15} dataset as a more realistic benchmark since it includes images of higher resolution and it spans $1000$ object classes.
	
	We evaluate the ability of our models to estimate the ground truth chroma channels $I_C$ given the intensity channel $I_G$. To quantify reconstruction quality of the estimated chroma channels we measure the mean squared error (MSE) on the chroma space. Since the end objective is to get good quality output images in the RGB space, we also evaluate the PSNR metric (measured in dB) of our estimated color image on the RGB output space. For decisions pertaining to encoding, we also evaluate performance with respect to the MS-SSIM~\cite{msssim} metric which is known to have strong correlations with human perception of quality. Note that MSE is measuring {\em error}, thus the lower the better. Instead PSNR and MS-SSIM measure quality, thus the higher the better.
	
	Tables~\ref{tb:archCIFAR} and\difftxt{~\ref{tb:archIN}} outline the architectures that were used to train our proposed model for estimating colors on the two datasets and also the details on how learning was performed. 	
	
	\subsection{\bf Zero-Cost Colorization}

	We begin by assessing different design choices and hyperparameters of our model in the context of zero-cost colorization.

	\paragraph{\difftxt{Branching Factor}}
We start by studying the impact of the branching factor in the context of a single color prediction (zero-cost colorization). To produce a single hypothesis from the multiple branches, here we use our trained branch predictor. On one hand, we expect that increasing the branching factor will allow our colorization model to better fit multi-modal color hypotheses. But on the other hand having more branches makes life harder for the predictor which needs to select the best branch for every pixel to produce the final output. This interpretation is confirmed by the results listed in Table~\ref{tb:branching_performance}, which reports colorization performance of our models on the CIFAR100 testing set. The ``Branch prediction accuracy'' reports the percentage of times the method chooses the branch yielding the minimum error with respect to the ground truth. It can be seen that when using only 1 branch the performance is poor despite no ambiguity in the selection of the branch. This model underfits the data, as it cannot represent multiple color modes. Conversely, a model with high branching factor (e.g., 7) does poorly because of the challenge posed by selecting the correct branch out of many. The best performance is obtained for a balanced trade-off between these two risks, achieved with an architecture of $5$ branches. 

\begin{table}[hb]
  \caption{Understanding the role of branching for zero-cost colorization on the CIFAR100 dataset. Accuracy measures the percentage of correct branch predictions made. The 3,5,7 Branch networks are trained with the branch predictor.}  
  \centering  
 \begin{tabular}{lcccc}  \label{tb:branching_performance}
%    \cmidrule{1-2}
    {\bf Architectures}     & MSE & PSNR & Branch prediction  \\
    						   &     &       & accuracy \\
    \midrule
    1 Branch 				  & 219.02 	& 23.72  & 100 \\
    3 Branches & 213.62  & 23.68  &  54.61\\
    5 Branches & 210.85  & 23.94  &  56.62\\
    7 Branches & 214.63  & 23.13  &  36.96\\        
  \end{tabular} 
\end{table}

\paragraph{\difftxt{Learning from Scratch vs Fine-Tuning}}
Colorization is a task that requires high level reasoning. While the results in Table~\ref{tb:branching_performance} were obtained by training the network from scratch for the purpose of colorization, it is interesting to study whether performance can be boosted by fine tuning the model from a deep network trained for high level reasoning tasks (e.g., image classification). 

Table~\ref{tb:finetuning_vs_lfs} shows results on both CIFAR100 and ImageNet when learning from scratch (LFS) as well as when fine-tuning (FT) the weights of our model from a network pretrained on image categorization (the  details of the categorization pretraining procedure are given in the appendix in Tables~\ref{tb:archCIFAR} and ~\ref{tb:archIN}). When fine-tuning we initialize the trunk with the weight of the pretrained categorization model. Then we add to the trunk layers that perform upsampling for dense prediction and the $K$ branches. The upsampling layers and the branches are initialized with random weights. The results in Table~\ref{tb:finetuning_vs_lfs} suggest 
that on CIFAR100 learning from scratch yields slightly better results than fine-tuning. Instead, on ImageNet fine-tuning the colorization network from a pre-trained categorization model is highly beneficial. We believe this happens because ImageNet pictures are much more complex then CIFAR100 photos. ImageNet pictures typically include one or more objects on more cluttered background scenes, whereas many of the CIFAR100 photos include only one object that fills the entire view. Thus, it makes sense that in the more complex setting of ImageNet, the colorization network benefits from the categorization pre-training for localizing objects and distinguishing them from the background. 
	
\begin{table}[ht]
  \caption{Studying the benefit of learning from scratch (LFS) vs fine-tuning (FT) a pre-trained categorization architecture for the purpose of zero-cost colorization.} 
  \centering
\begin{tabular}{lllllll}\label{tb:finetuning_vs_lfs}
%    \cmidrule{1-2}
    {\bf Architectures}     & MSE & PSNR \\
	\midrule
	CIFAR100 \\
%    \midrule
%    1 Branch-LFS 				  & 219.02 	&  23.72   \\
%    1 Branch-FT 					  & 	192.44 	&  24.19 	\\ 
    \midrule      
    5 Branch-LFS 				  & 210.85  	&  23.94   	\\
    5 Branch-FT 					  & 250.28  &  23.64 	\\   
    \midrule
    \midrule
    \difftxt{ImageNet} \\
    \midrule
    5 Branch-LFS					  & 232.64 & 23.13   \\
    5 Branch-FT					  & 194.12 & 23.72   \\ 	
    \midrule        	
  \end{tabular}
\end{table}	

\paragraph{\difftxt{Analysis of learned color models}}
\difftxt{By visually inspecting the multiple hypotheses generated by our network in Table~\ref{tb:diverseoutputs} we can observe that the branches specialize well in their ability to represent different color modes. Branch 3 tends to colorize grass and bushes as brown (a plausible color) whereas Branch 2 colorizes them as green. Both branches are implicitly able to recognize and localize grass and plants and propose appropriate colors for them without ever being explicitly taught the notion of grass. The table also shows us that our network learns a compositional method for recreating the colors of objects that are highly multi-modal with various branches specializing in dominantly observed colors. Branch 5 predicts red and shades of red for objects that exhibit numerous colors and Branch 4 predicts blue as these colors dominate the data used to train our models.}	

\setlength{\tabcolsep}{1pt}
\begin{table*}[t!]
\centering
{\small
\begin{tabular}[ht!]{c|ccccc|c}
%\centering
\hline
Original &  Dahl  & Zhang  & Iizuka  & Larsson & Ours & Ours \\
&  \cite{dahl2015}  & et al.~\cite{zhang2016colorful} & et al.~\cite{IizukaSIGGRAPH2016} & et al.~\cite{larsonLearningRepresentations} & zero-cost & low-cost \\
\midrule
\includegraphics[width=2.2cm]{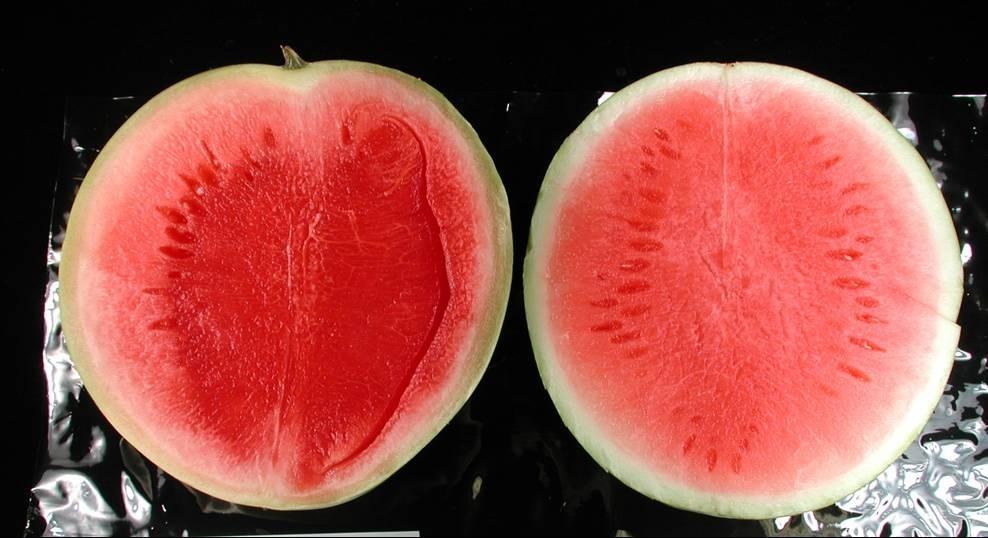} & \includegraphics[width=2.2cm]{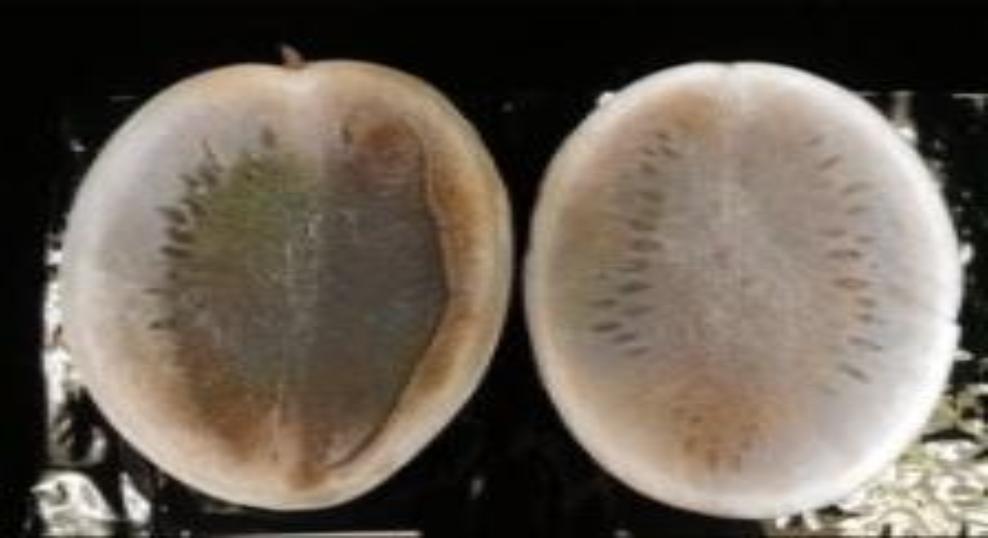}& \includegraphics[width=2.2cm]{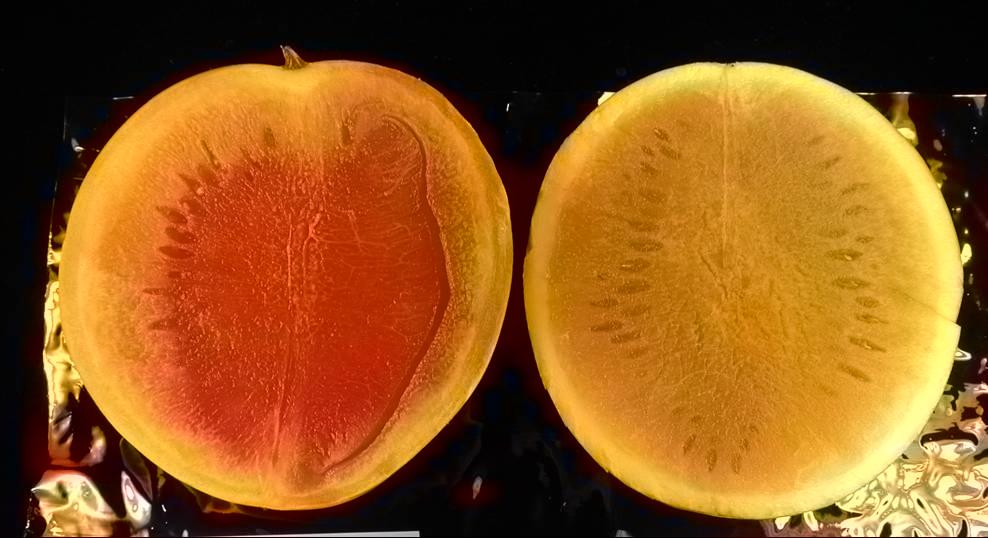}& 
\includegraphics[width=2.2cm]{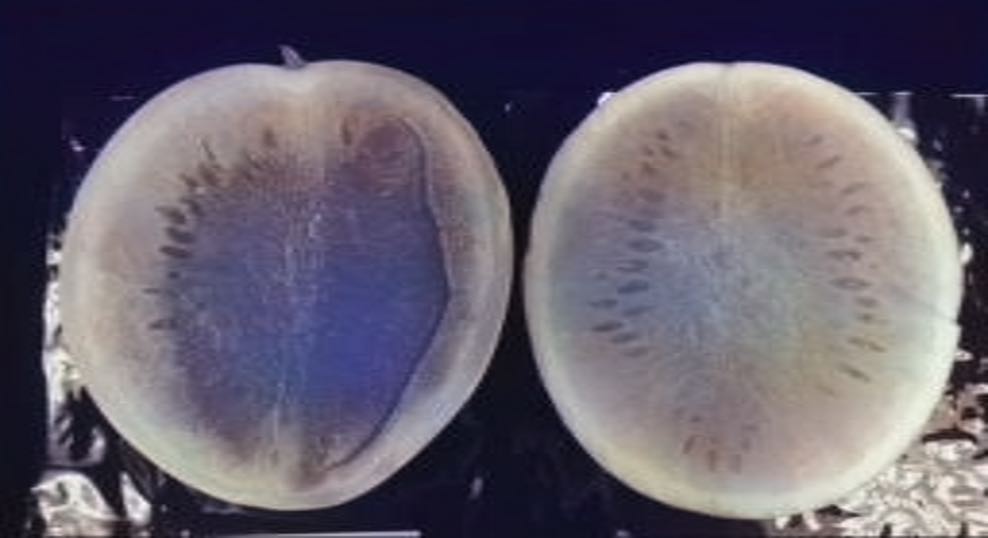}& 
\includegraphics[width=2.2cm]{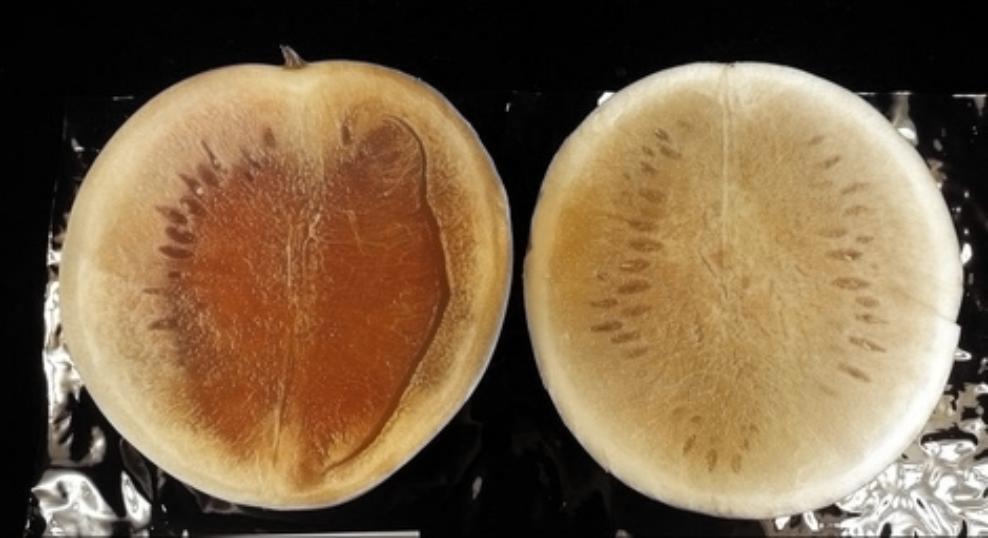}& 
\includegraphics[width=2.2cm]{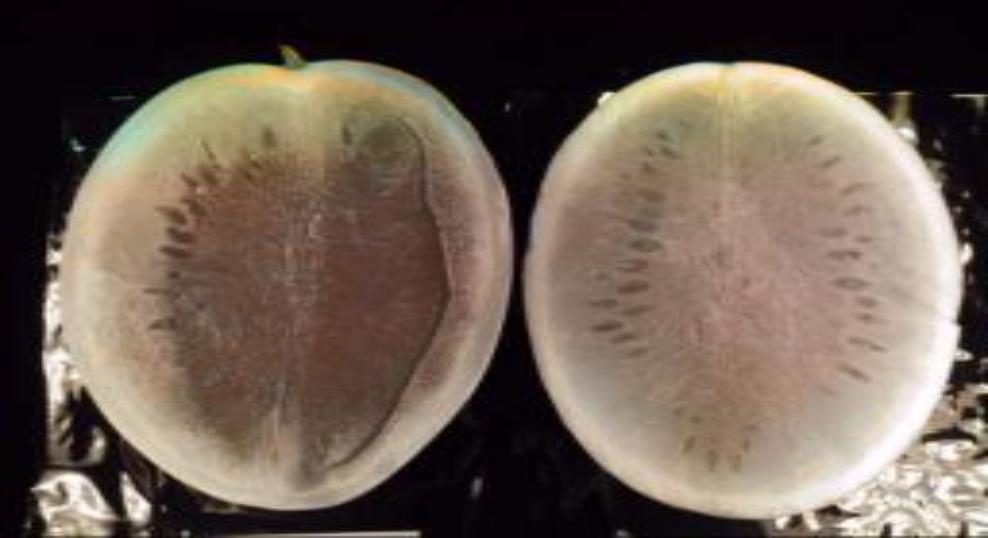}&  \includegraphics[width=2.2cm]{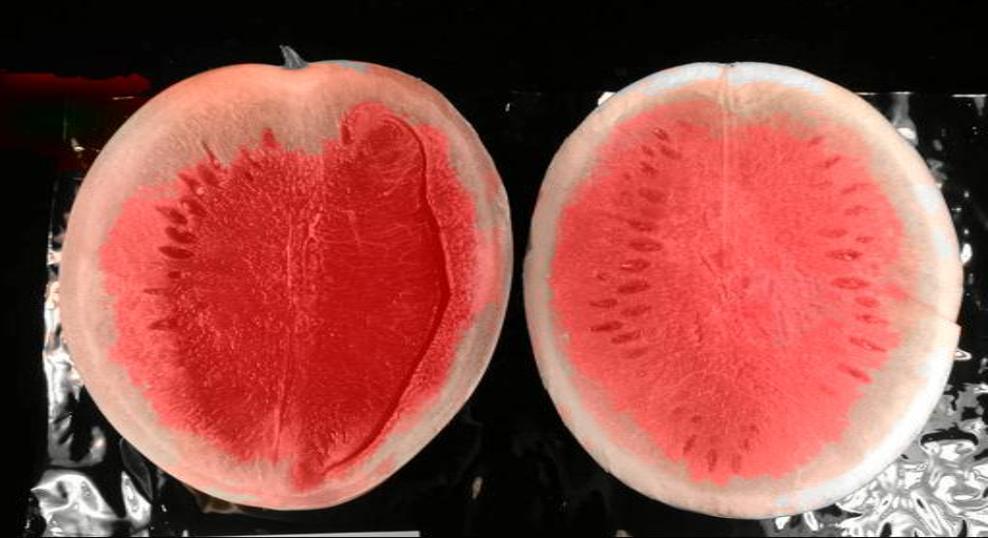}\\
\includegraphics[width=2.2cm]{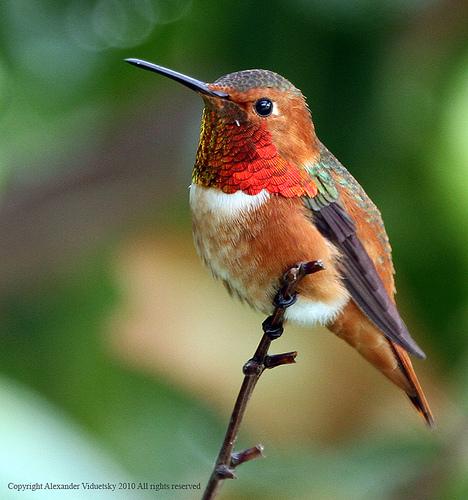} & \includegraphics[width=2.2cm]{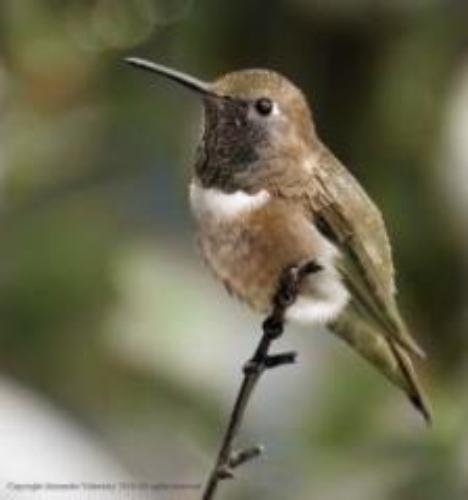}& \includegraphics[width=2.2cm]{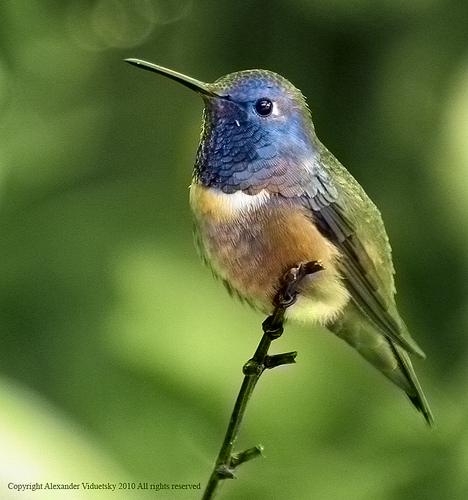}& 
\includegraphics[width=2.2cm]{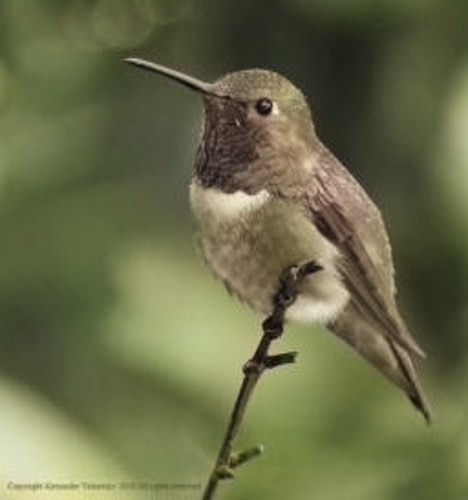}& 
\includegraphics[width=2.2cm]{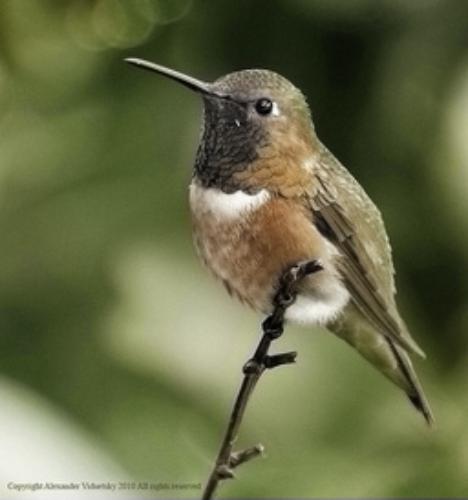}& 
\includegraphics[width=2.2cm]{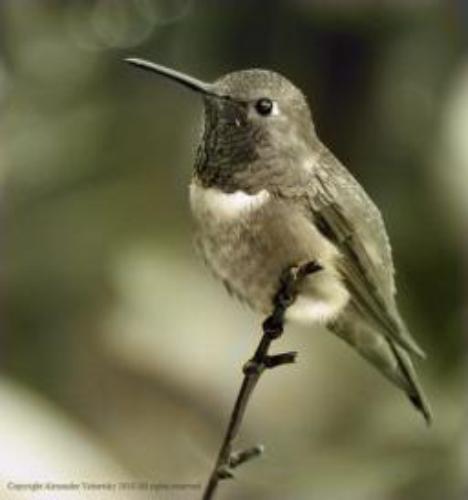}&  \includegraphics[width=2.2cm]{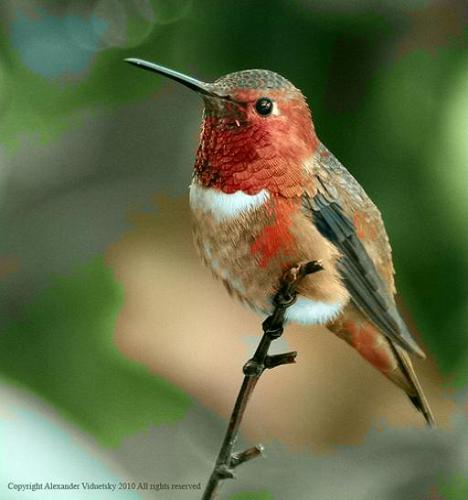}\\
\includegraphics[width=2.2cm]{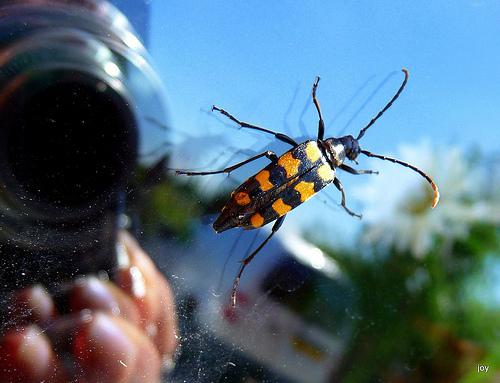} & \includegraphics[width=2.2cm]{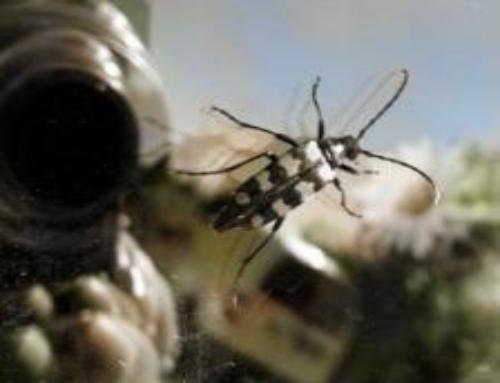}& \includegraphics[width=2.2cm]{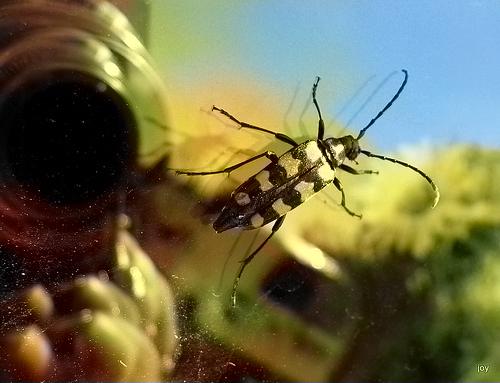}& \includegraphics[width=2.2cm]{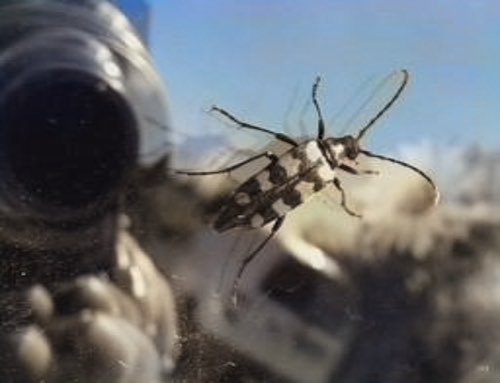}&  \includegraphics[width=2.2cm]{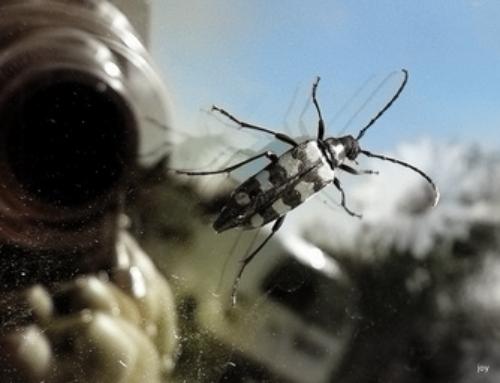}& 
\includegraphics[width=2.2cm]{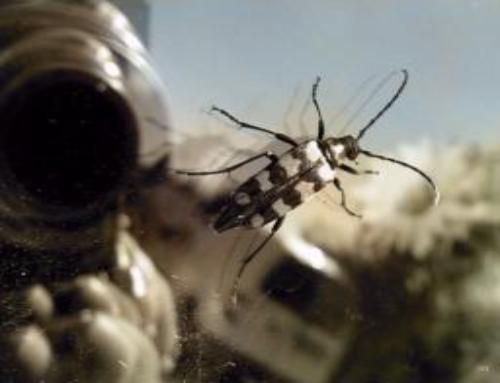}&  \includegraphics[width=2.2cm]{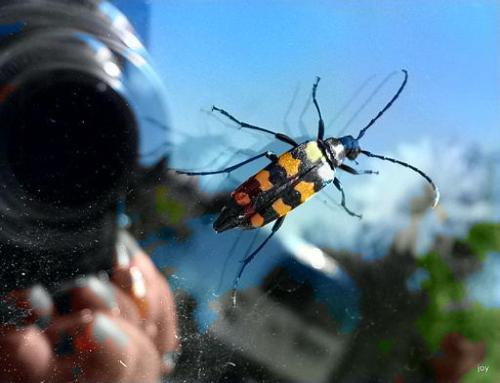}\\
\includegraphics[width=2.2cm]{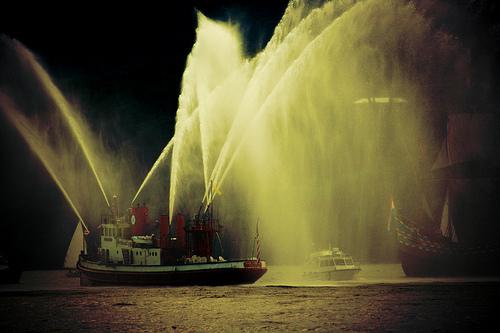} & \includegraphics[width=2.2cm]{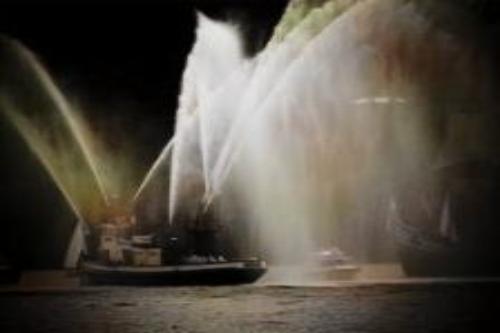}& \includegraphics[width=2.2cm]{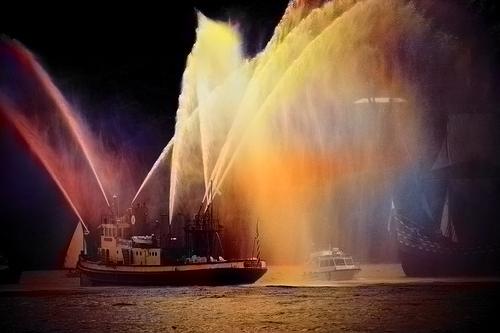}& \includegraphics[width=2.2cm]{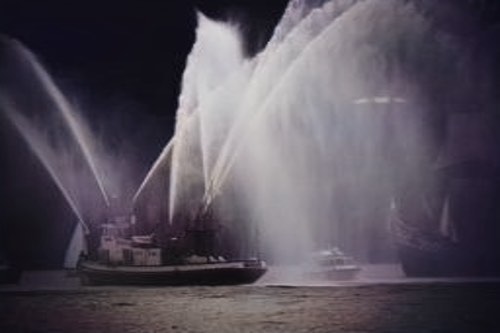}&  \includegraphics[width=2.2cm]{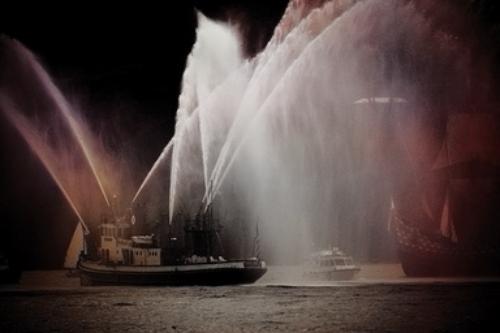}& 
\includegraphics[width=2.2cm]{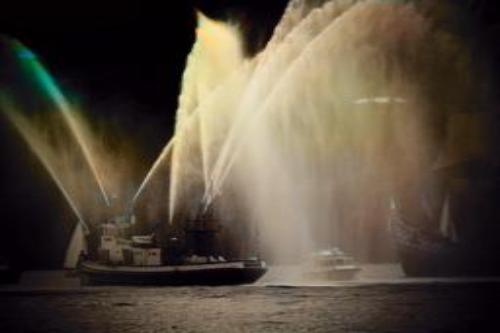}&  \includegraphics[width=2.2cm]{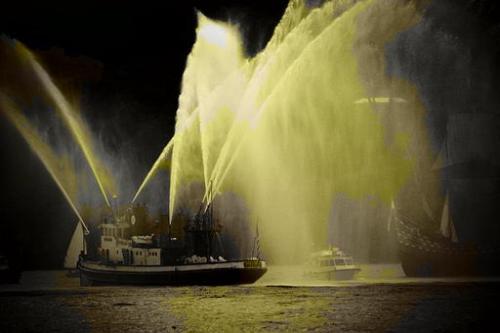}\\
\includegraphics[width=2.2cm]{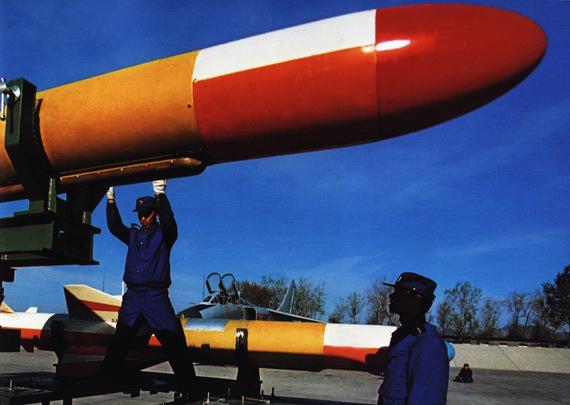} & \includegraphics[width=2.2cm]{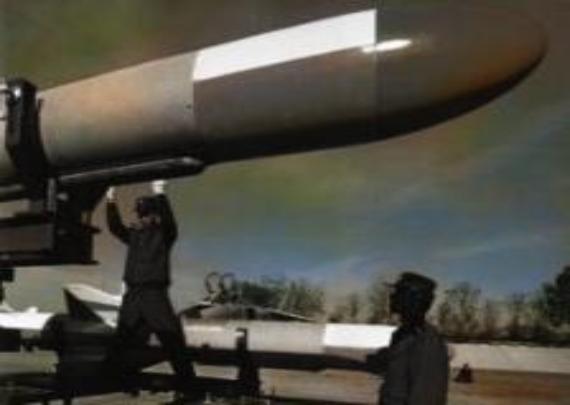}& \includegraphics[width=2.2cm]{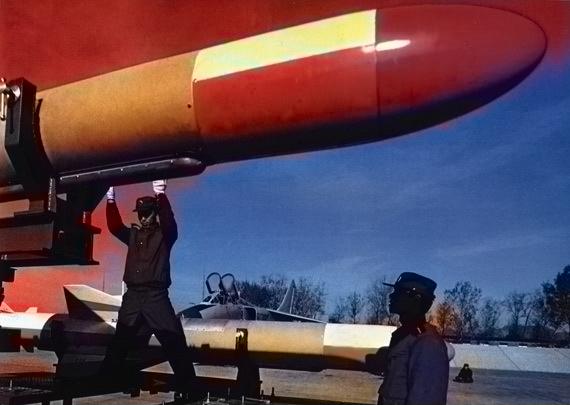}& \includegraphics[width=2.2cm]{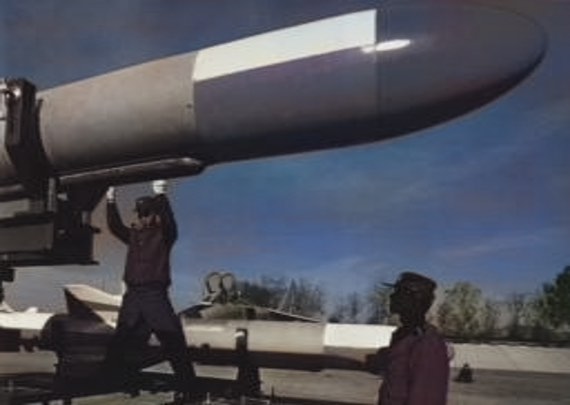}&  \includegraphics[width=2.2cm]{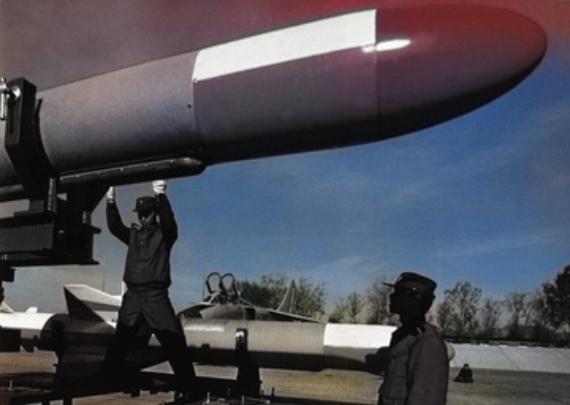}& 
\includegraphics[width=2.2cm]{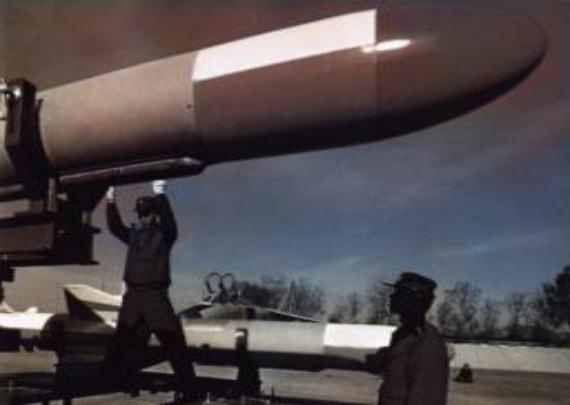}&  \includegraphics[width=2.2cm]{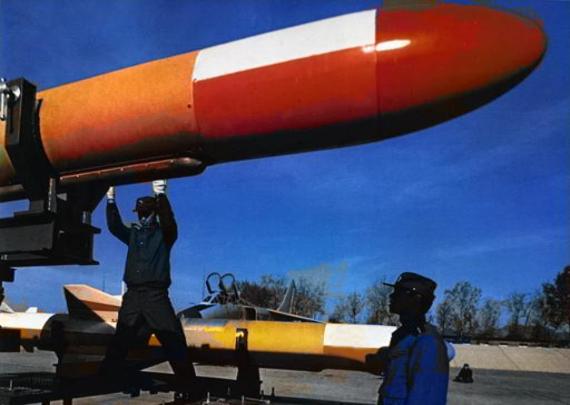}\\
\includegraphics[width=2.2cm]{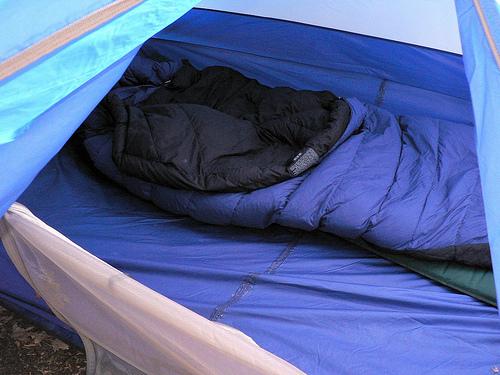} & \includegraphics[width=2.2cm]{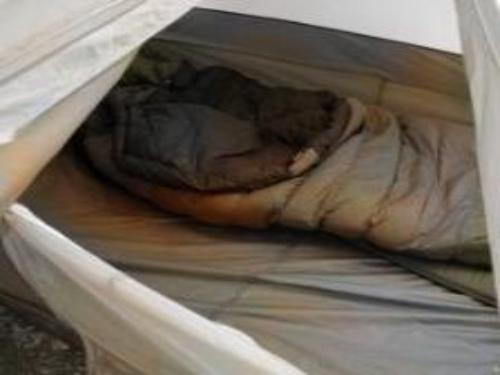}& \includegraphics[width=2.2cm]{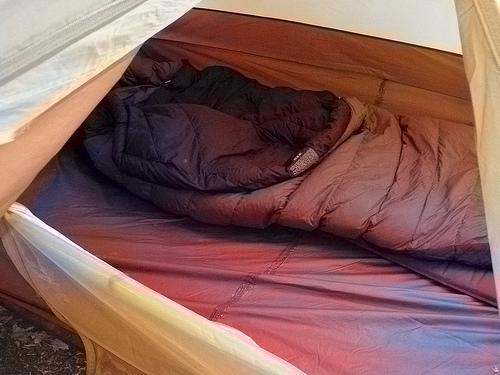}& \includegraphics[width=2.2cm]{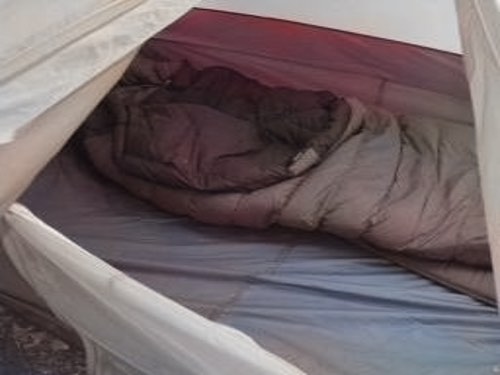}&  \includegraphics[width=2.2cm]{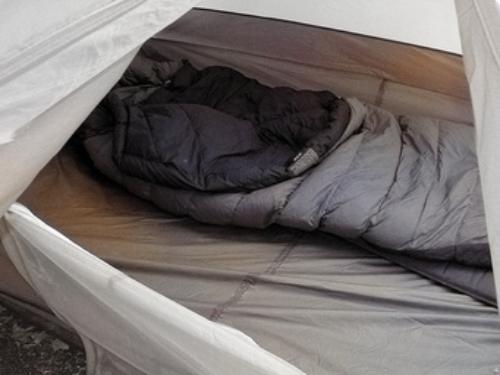}& 
\includegraphics[width=2.2cm]{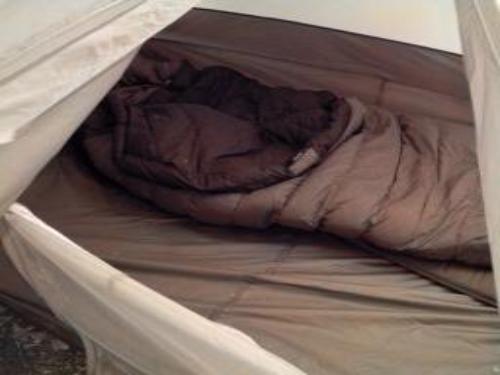}&  \includegraphics[width=2.2cm]{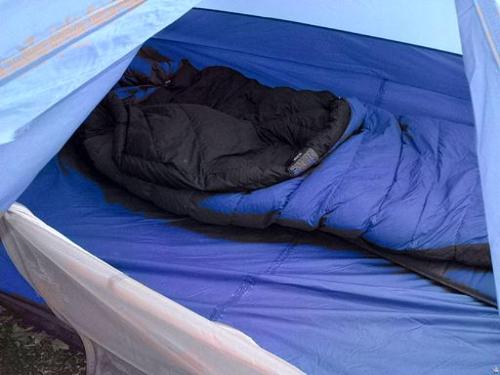}\\
\includegraphics[width=2.2cm]{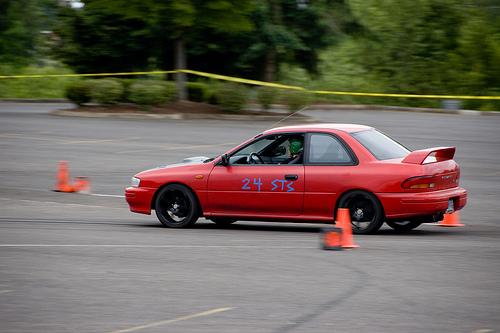} & \includegraphics[width=2.2cm]{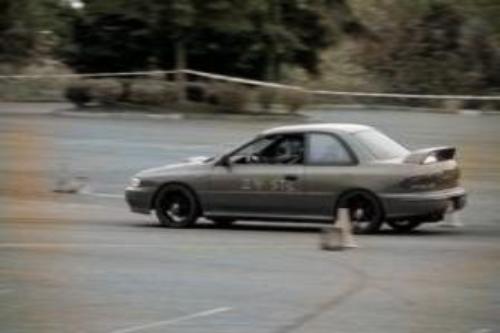}& \includegraphics[width=2.2cm]{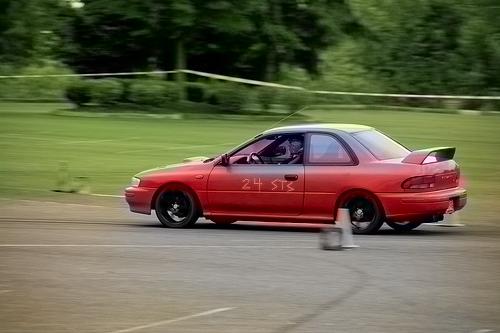}& \includegraphics[width=2.2cm]{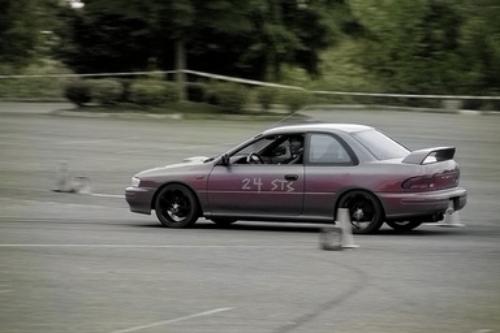}& \includegraphics[width=2.2cm]{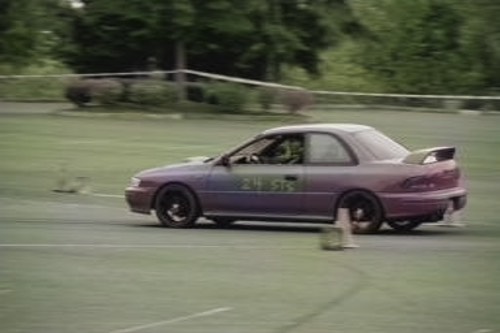}& 
\includegraphics[width=2.2cm]{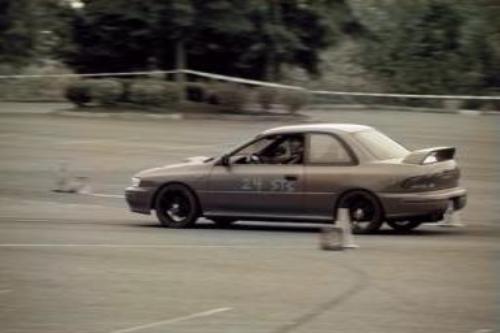}&  \includegraphics[width=2.2cm]{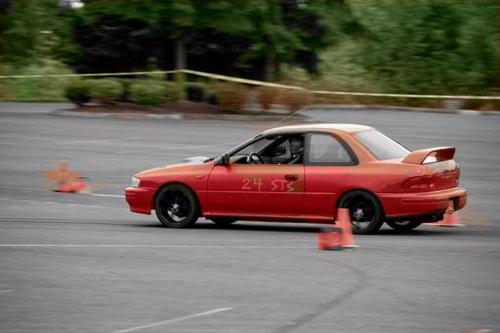}\\
\includegraphics[width=2.2cm]{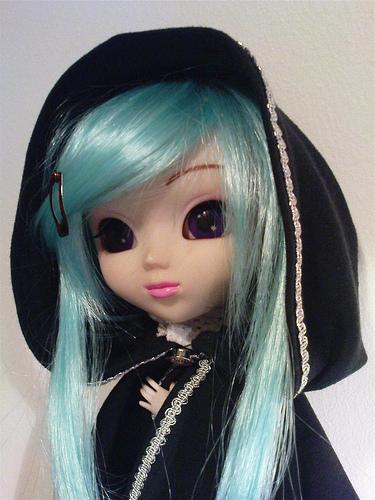} & \includegraphics[width=2.2cm]{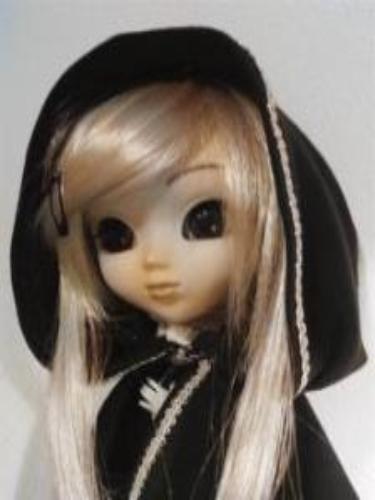}& \includegraphics[width=2.2cm]{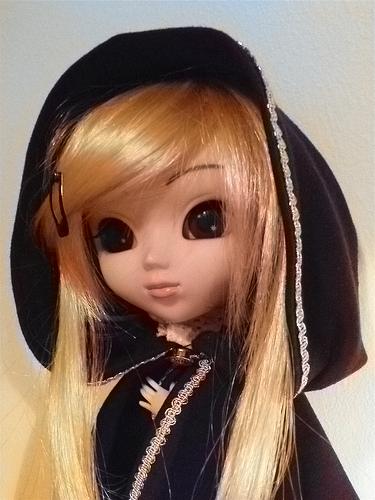}& \includegraphics[width=2.2cm]{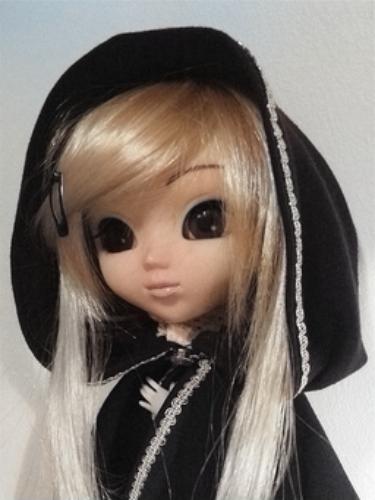}& \includegraphics[width=2.2cm]{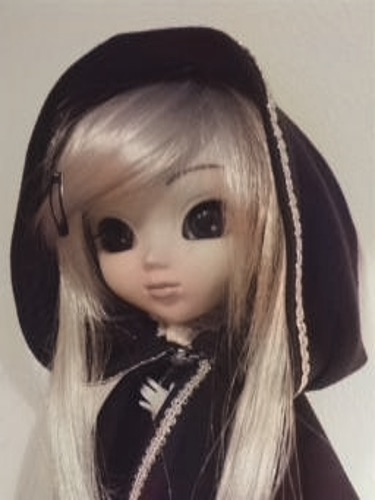}& 
\includegraphics[width=2.2cm]{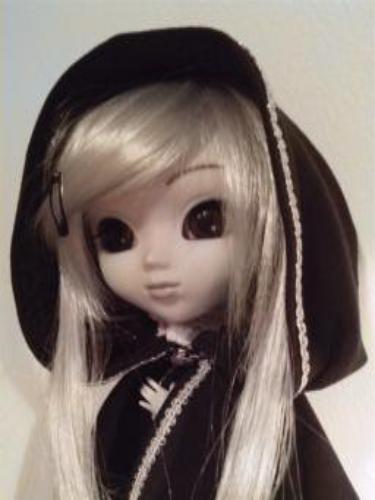}&  \includegraphics[width=2.2cm]{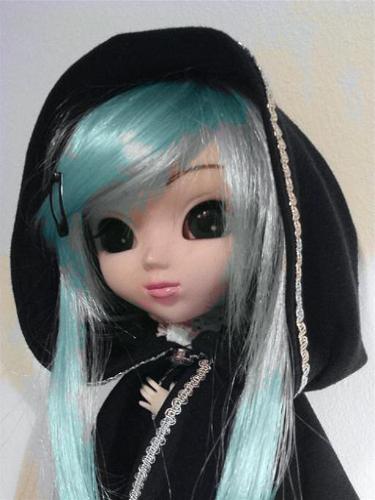}\\
\includegraphics[width=2.2cm]{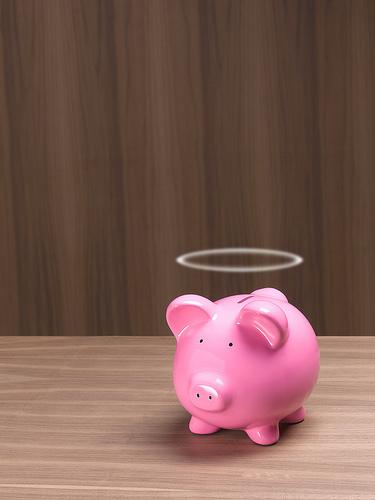} & \includegraphics[width=2.2cm]{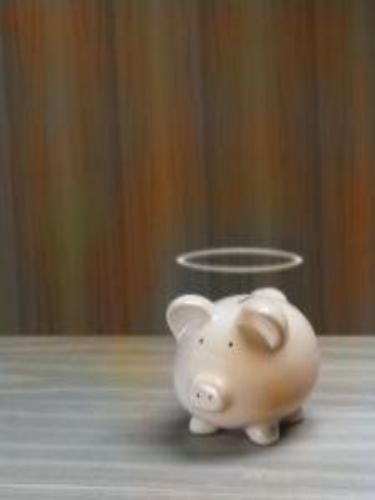}& \includegraphics[width=2.2cm]{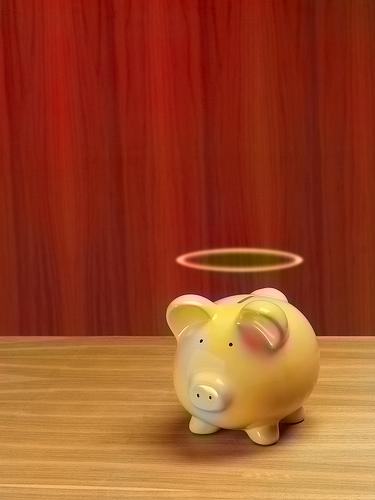}& \includegraphics[width=2.2cm]{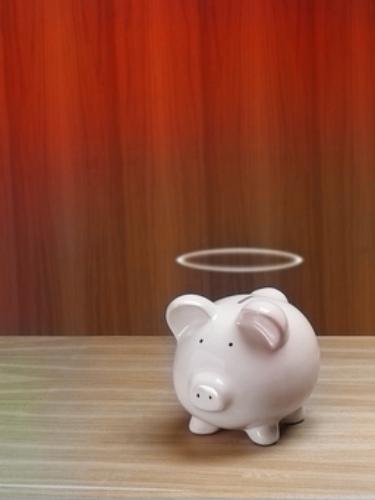}& \includegraphics[width=2.2cm]{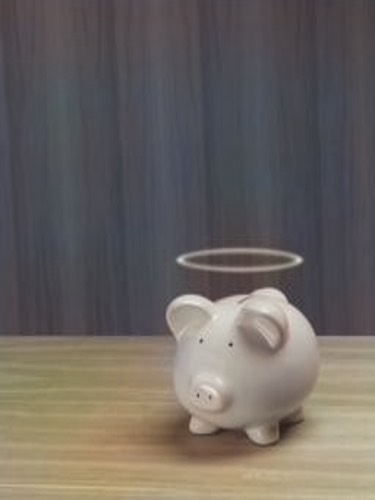}& 
\includegraphics[width=2.2cm]{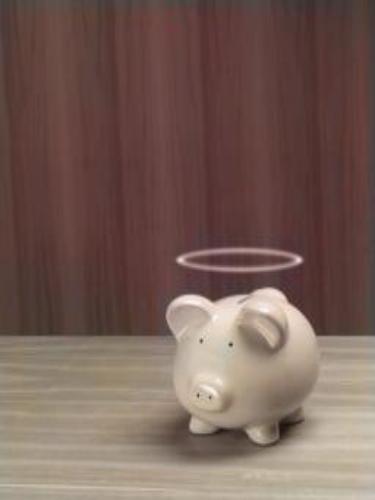}&  \includegraphics[width=2.2cm]{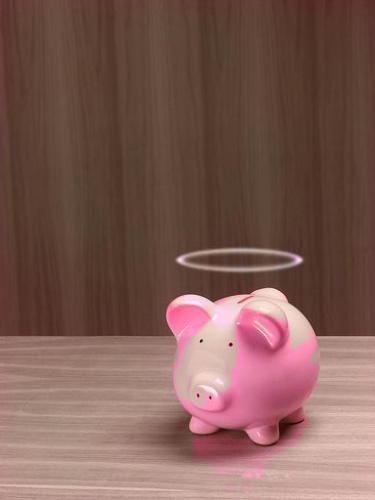}\\
\end{tabular}
}
\caption{Qualitative comparison of results produced by several recently proposed deep architectures for image colorization as well as our zero-cost and \difftxt{low-cost} method.}
\label{tb:comparingApproaches}
\end{table*}

\paragraph{\difftxt{Comparison with the State-of-the-Art in Zero-Cost Colorization}}

We now perform a quantitative and qualitative comparison of our approach against various recently introduced methods for zero-cost colorization utilizing the large-scale ImageNet dataset. The training is performed on the \LT{training split of the} \difftxt{ImageNet} dataset and testing is done on the \difftxt{ImageNet}-10k benchmark introduced by \difftxt{Larsson} et al..

Table~\ref{tb:imagenet_benchmark} summarizes the quantitative results on the zero-cost colorization task. The first two rows represent weak baselines intended to help us understand the difficulty of the colorization task on this benchmark. The first row shows performance obtained by predicting the constant average chroma irrespective of the input. The average is calculated per-pixel over the entire training set. The second row makes use of ground-truth class labels and outputs for each test image the per-pixels color average computed from the training examples belonging to the class of the input image. This baseline is useful as the colors in the image are strongly correlated to the category information.

Based on our results in the previous two subsections, for this experiment we used a network with $K=5$ branches fine-tuned from the pre-trained categorization model of He et al.~\cite{deepresidualHe2015}. We can observe that, although zero-cost colorization is not the focus of our work, our model with branch prediction outperforms the recent approaches of Dahl~\cite{dahl2015} and Zhang et al.~\cite{zhang2016colorful}, which were designed for this task. Our method is second behind \difftxt{Larsson} et al.~\cite{larsonLearningRepresentations}.

\begin{table}[htb]
  \caption{Evaluation of various approaches for zero-cost colorization on the \difftxt{ImageNet}-10k benchmark proposed by \difftxt{Larsson} et al.~\cite{larsonLearningRepresentations}. The table measures the performance of the networks in their ability to reproduce the original color image from the grayscale version.}
  \label{tb:imagenet_benchmark}
  \centering
  \begin{tabular}{lllll}
    \toprule
%    \cmidrule{1-2}
 	Architectures     & MSE & PSNR \\
    \midrule
  Per-Pixel Average Chroma Over all Classes	& 225.34	 & 22.78 \\
  Class-Specific Per-Pixel Average Chroma & 198.70 & 23.26 \\
    \midrule    
   		 Dahl \cite{dahl2015} 		&  	199.17 	&   22.88    \\        
	 \difftxt{Larsson} et al. \cite{larsonLearningRepresentations}	& 	154.69	&	24.80 	\\
	Zhang et al. \cite{zhang2016colorful}	& 	270.17 & 21.58 	\\
    \midrule	
Ours - 5 branches w/ branch predictor & 194.12 & 23.72 \\ 
    \bottomrule
  \end{tabular}
\end{table}

We now turn to a qualitative analysis of our results. In \difftxt{Table}~\ref{tb:comparingApproaches} we show how different approaches perform when tasked with colorizing certain images. We can see that our zero-cost colorization with branch prediction is able to generate accurate colors for objects whose chroma is unambiguous (e.g., grass or road). In some cases it can produce {\em plausible} colors for objects that have multiple possible colors, although the plausible output may differ from the ground truth. In the last column of Table~\ref{tb:comparingApproaches}  we show the output produced by our low-cost encoding scheme \LT{(using the oracle encoded with the lossy SLIC superpixel compression)}, which is able to approximate with high-fidelity the true colors for all objects contained in the example images. This illustrates the importance of adopting a non-zero-cost approach when the intended objective is compression.

\subsection{\bf Low-Cost Colorization for Image Compression}

In this subsection we study our approach in the regime of low-cost colorization for image compression.

\begin{figure}[ht]
\centering
	  \includegraphics[width=8.0cm]{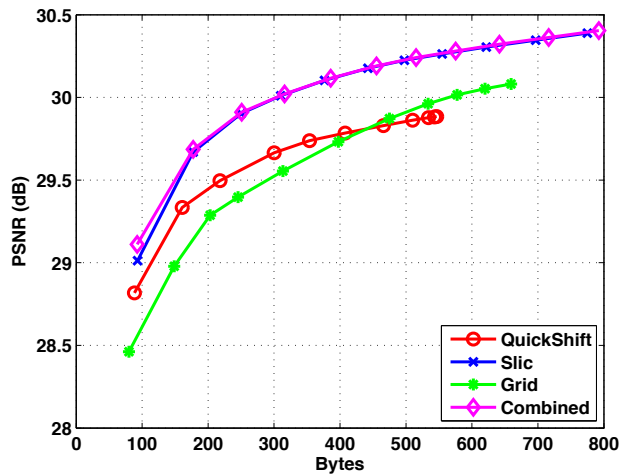}
	  \caption{We show how various methods for encoding the oracle in a lossy fashion perform with respect to each other. The combined curve shows the maximum benefit that can be achieved by using these three approaches in conjunction.}\label{fig:decisionRegion}
\end{figure}

\paragraph{\bf Lossy Encoding of the Oracle}
	In order to reduce storage space, we proposed to perform region-based approximations of the oracle encoding. Figure~\ref{fig:decisionRegion} shows PSNR achieved for three different methods for defining regions: SLIC~\cite{slic}, QuickShift~\cite{quickshift} and a regular grid. The first two are segmentation methods 
that take as input the intensity image $I_G^{(i)}$ and subdivide it into segments according to low-level cues (similar grayscale intensity values in nearby pixels). \difftxt{The} figure also shows the impact of varying the size of the regions. For the segmentation methods, the number of regions is varied by adjusting hyperparameters (region size and spatial regularization for SLIC and ratio of intensity to spatial localization, kernel size for computing similarity and maximum distance to search in for neighbors with Quickshift). We can see that SLIC tends to performs the best out of the the 3 proposed methods. As we are performing compression, storing which of these three methods to use requires fairly low-cost. Thus, we also include in the figure the method \difftxt{``Combined"} which selects the best of the three strategies for a given encoding size. We note that  encoding the oracle without the region-based approximations would require a storage space of 4490 bytes. The oracle yields a PSNR of 30.89 (dB)\difftxt{.} Thus, overall the region-based encodings provide orders of magnitude \difftxt{more} savings in space at very little loss in quality. \difftxt{We also note that the global correction produces an improvement \LT{$1.83$} (dB) on average. The improvement produced by global correction varies based on the number of superpixels being used to encode the oracle decisions. Improvements are smaller when fewer superpixels (of larger sizes) are used as many pixels (within each superpixel) are being incorrectly assigned to the wrong branch. As the number of superpixels used increases the improvements brought about by global correction increase. The improvement from global correction becomes fairly constant as we approach near pixel level segmentation.}
	
\paragraph{\bf \difftxt{Trading off} Size vs Quality for Image Compression}

	In this section we consider several schemes of colorization for image compression and evaluate their ability to generate different quality levels by varying the storage size. Training was done on the entire \difftxt{ImageNet} dataset and testing was performed on the \difftxt{ImageNet}-10k benchmark introduced by \difftxt{Larsson} et al. We include in this analysis both zero-cost and low-cost approaches. Note that while He et al.~\cite{he2009unified} and Cheng et al.~\cite{cheng2007learning} have also proposed methods for low-cost colorization, we are unable to include them in this quantitative comparison as neither code nor predictions have been released for these systems (but we present a {\em qualitative} comparison to these approaches in the next subsection).  We include JPEG compression applied to only the color channels. For our oracle encoding,  we used DPCM (Differential pulse-code modulation) followed by Huffman encoding~\cite{huffmanUsed}) in order to perform lossless compression. We also include results based on lossy encoding of the oracle using our image region decompositions (described in section~\ref{sec:lossOracle}). For all our models we used a 5-branch architecture fine tuned from the pretrained categorization network described in\difftxt{~\ref{App:AppendixA}}.
	
The curve denoted as ``ours: low-cost'' corresponds to selecting the best strategy from our three lossy encoding schemes (SLIC, QuickShift, Grid). The size corresponding to the encoding of an image includes the bits needed for storing which lossy method is being used, its associated hyperparameter settings for generating regions of a certain size and global correction parameters in addition to the branch index for each region. 
We observe that our method is by far the best in the extremely low file-size regime. For the same quality, it is several orders of magnitude more compact than JPEG.

\subsection{\bf Comparison to Other Colorization Approaches for Compression}

\setlength{\tabcolsep}{1pt}
\begin{table*}
\centering
\begin{tabular}[ht!]{c|ccccc|c}
\hline
Original Image & Cheng et al.~\cite{cheng2007learning} & He et al.~\cite{he2009unified} & Ours  \\
\midrule
& $6,524$ bytes & $6,524$ bytes & $512$ bytes\\
\includegraphics[height=2.3cm]{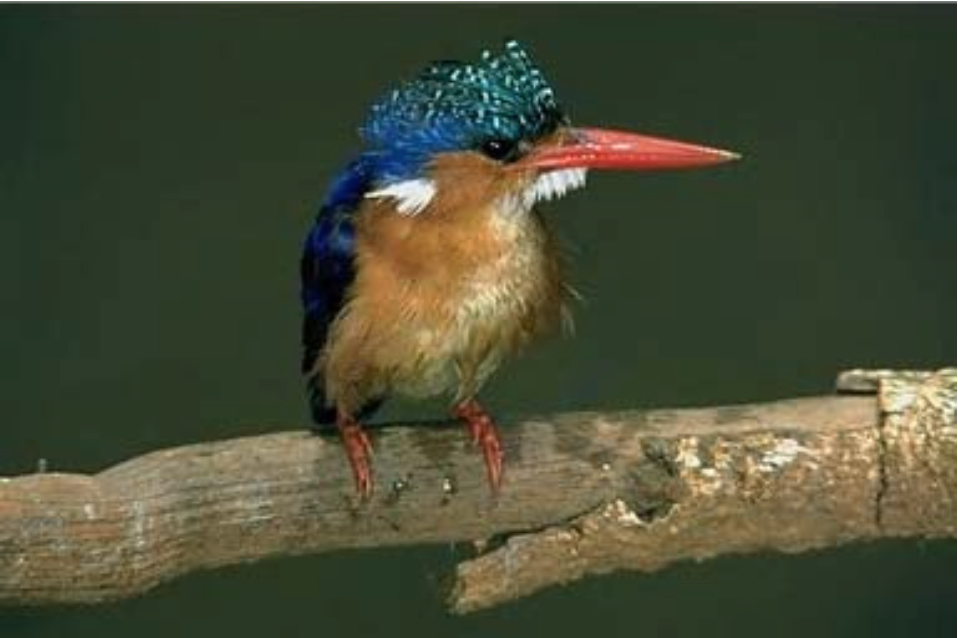} & \includegraphics[height=2.3cm]{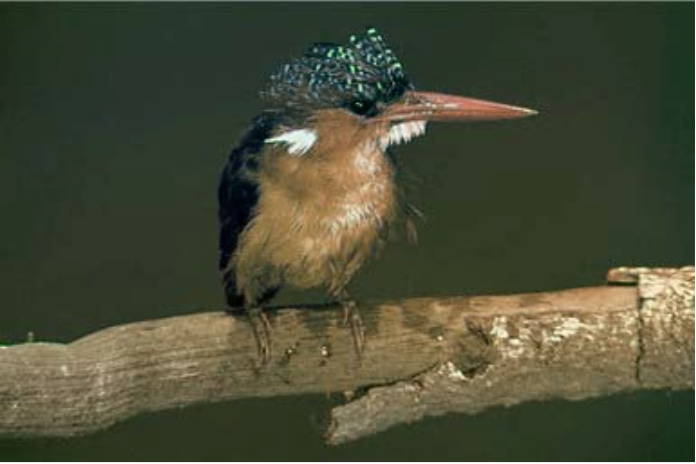}& \includegraphics[height=2.3cm]{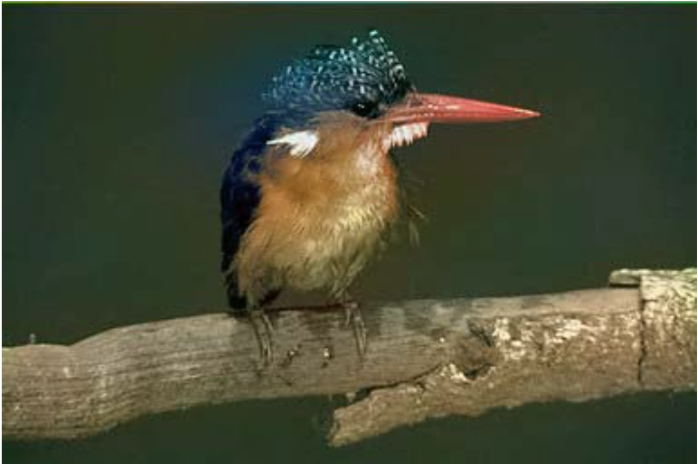}& 
\includegraphics[height=2.3cm]{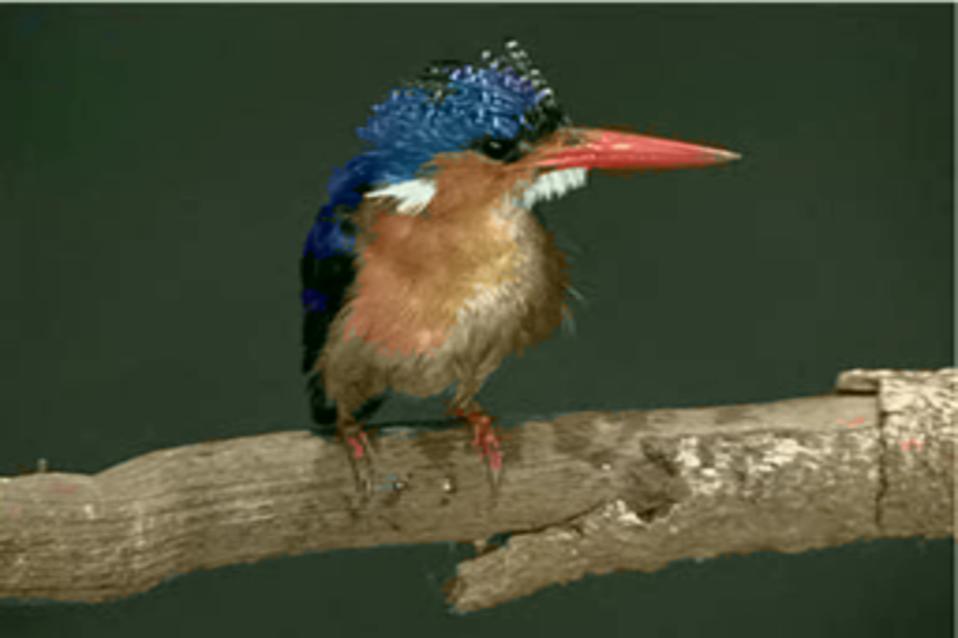}\\
\hline
 & $6,900$ bytes & $6,900$ bytes & $674$ bytes \\
\includegraphics[height=2.3cm]{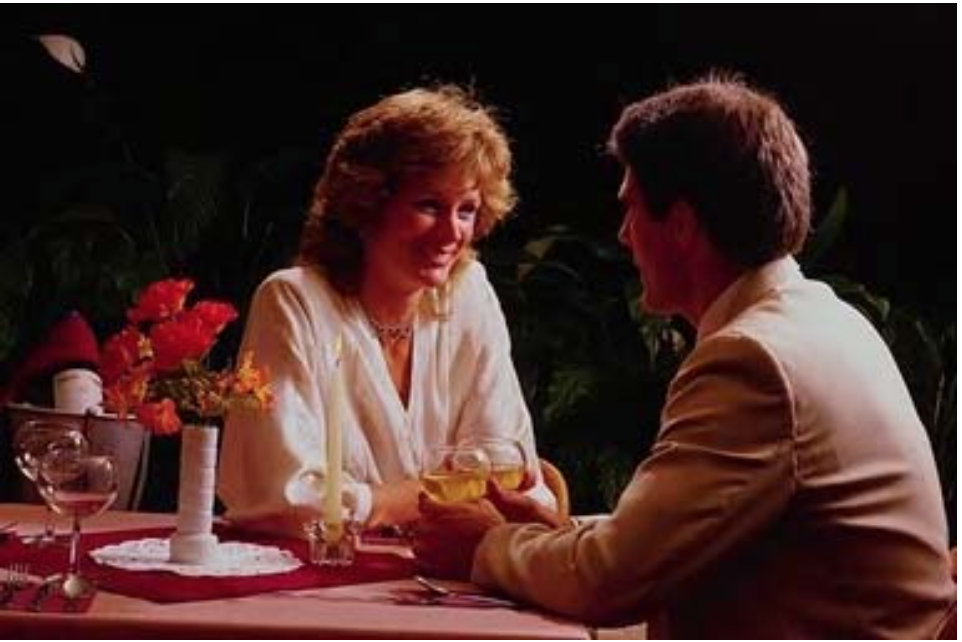} & \includegraphics[height=2.3cm]{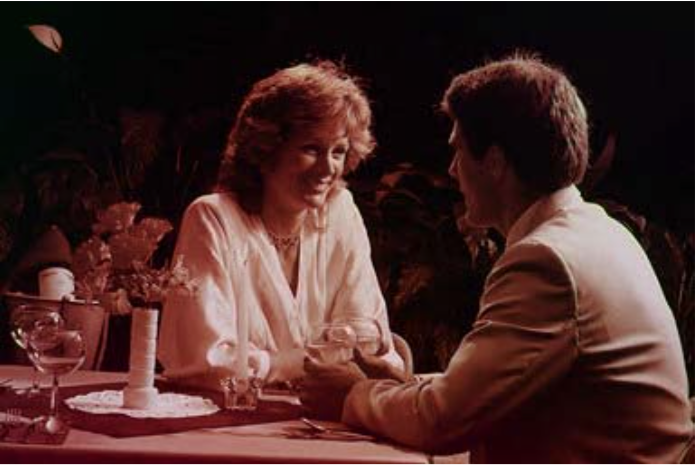}& \includegraphics[height=2.3cm]{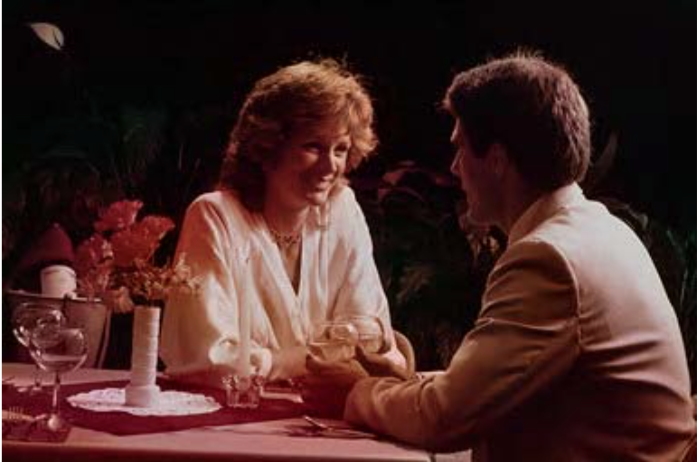}& \includegraphics[height=2.3cm]{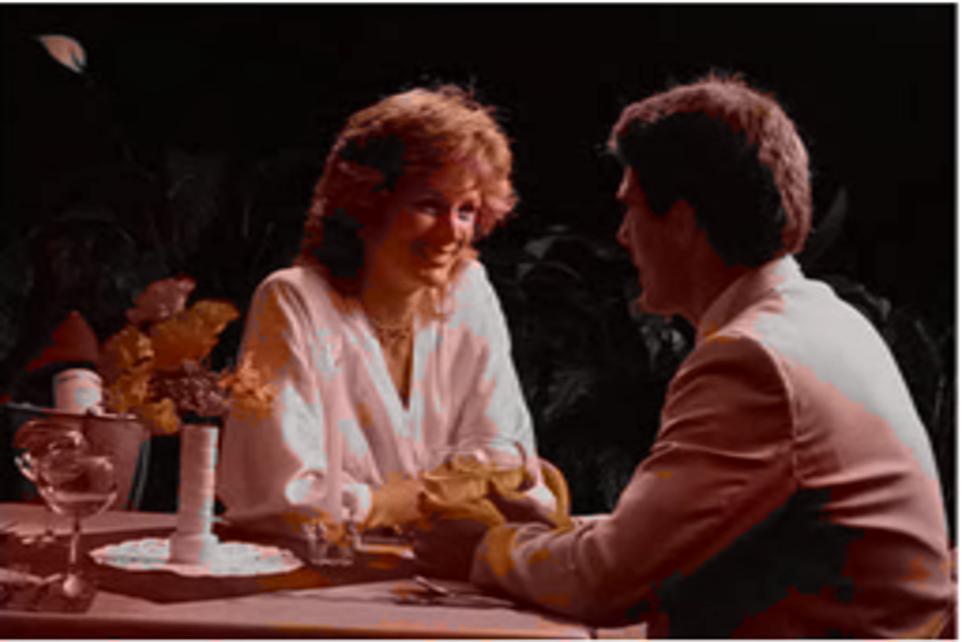}\\
\hline
 & $7,536$ bytes & $7,536$ bytes & $269$ bytes \\
\includegraphics[width=2.3cm]{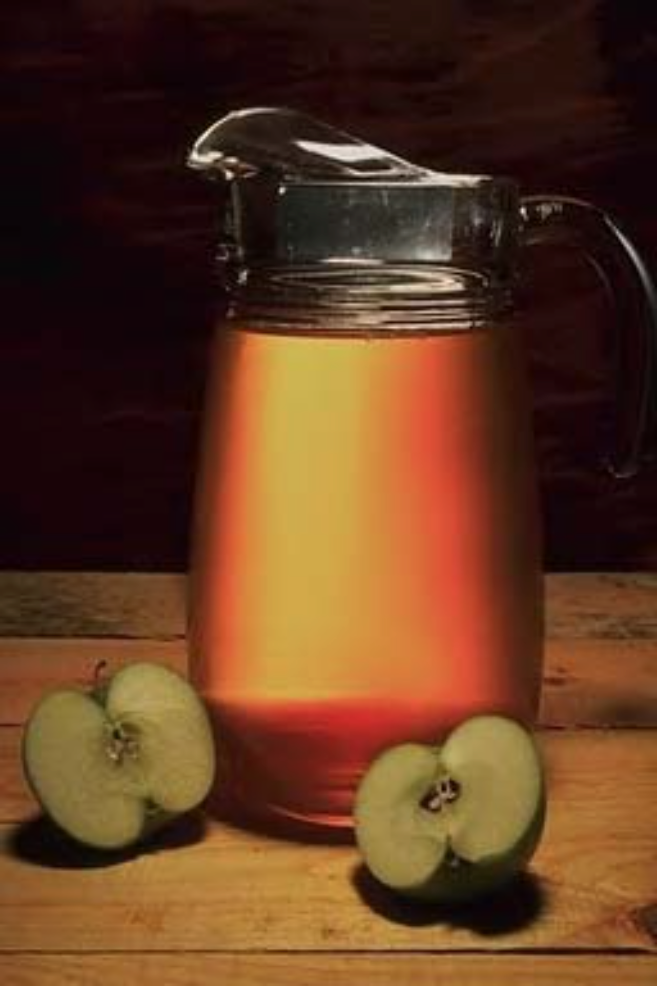} & \includegraphics[width=2.3cm]{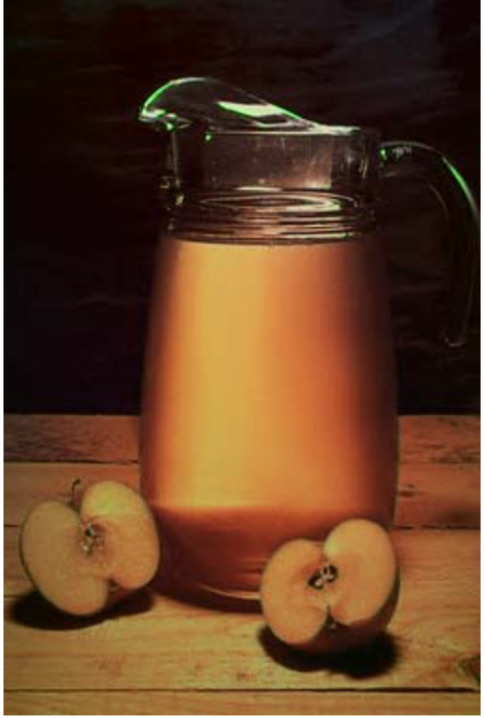}& \includegraphics[width=2.3cm]{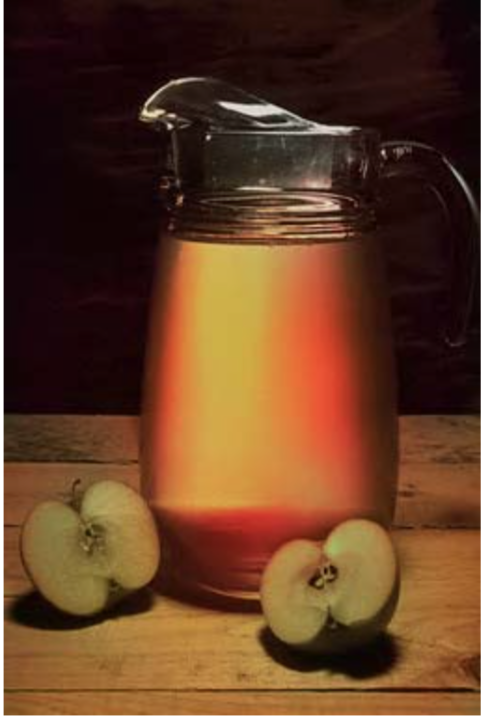}& 
\includegraphics[width=2.3cm]{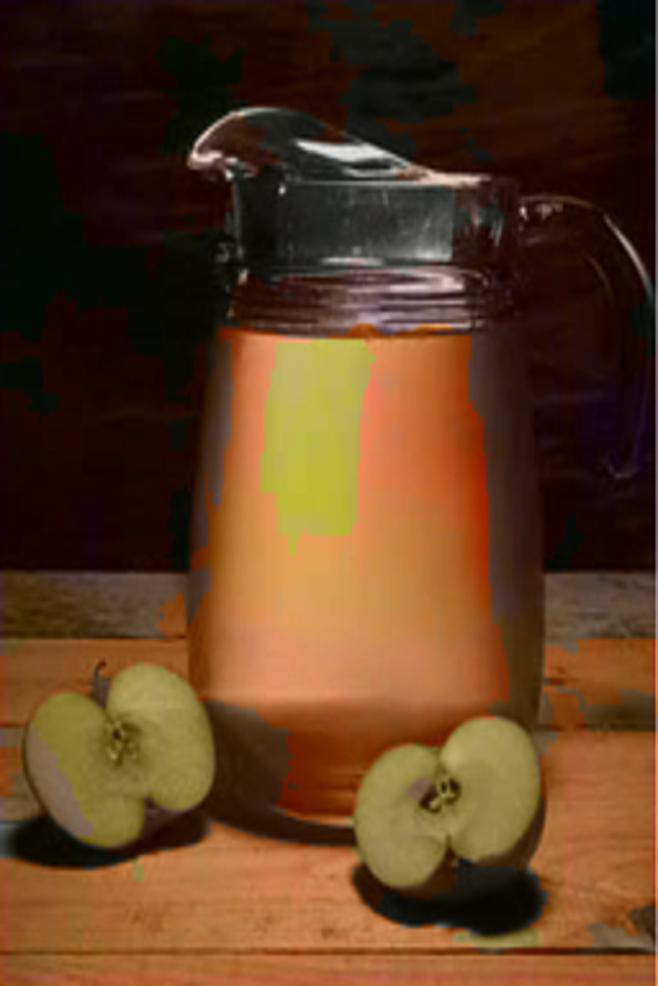} \\
\end{tabular}
\caption{Qualitative comparison between the colorization approaches for compression of He et al.~\cite{he2009unified}, Cheng et al.~\cite{cheng2007learning}  and our low-cost method.}
\label{tb:comparingApproaches2}
\end{table*}

\difftxt{Here we compare  our method with the approach of He et al.~\cite{he2009unified} and Cheng et al.~\cite{cheng2007learning}. Since no code or predictions for these methods are available, we perform a {\em qualitative} comparison on $3$ images taken from the paper of He et al.. The results are shown in Figure ~\ref{tb:comparingApproaches2}. Our method can produce faithful approximations of the original color images from their grayscale versions at a fraction of the storage size needed by the approaches of He et al. and Cheng et al.. We invite the reader to pay particular attention to the colors assigned to the bird feet: our method is the only one that can faithfully reproduce the original colors, despite having much lower storage footprint.}

\difftxt{We note that this is a qualitative study limited to only 3 images and as such is non-conclusive. However, in absence of the possibility to compare quantitatively our method to the these other approaches, we present these results to highlight certain differentiating characteristics of our technique with respect to prior work. First, our method is able to generate images that are significantly smaller in size at competitive quality. The high compression rate achieved by our approach derives from the fact that it merely requires storing a single value (the branch index) per-pixel (in case of the oracle) or per-region (in case of lossy oracle), while the actual chroma values are generated ``for free'' by the network. In contrast, the approaches of He et al. and Cheng et al. perform propagation of colors from the seed locations to all pixels and thus require storing the seed locations (two 8-bit numbers per seed) in addition to the chroma values at the given seeds (other two 8-bit numbers per seed). }

\difftxt{A second major advantage of our approach lies in the very fast decoding process. In comparison, the methods of He et al. and Cheng et al. need to solve a costly optimization for color propagation. We report the details of this computational cost comparison in the next subsection. }

\difftxt{Third, our approach offers the advantage of learning visual features which may be useful for other tasks. While we reserve to future work the performance investigation of how well our features transfer to other high-level tasks (e.g., categorization), we believe that our training procedure for colorization may in principle serve as a beneficial form of unsupervised pretraining to initialize deep networks in scenarios where labelled data is scarce.}

\difftxt{Finally, our system has the ability to be domain adaptive, since it is a model learned from data. By restricting the data to the application domain, the model can become more specific with potential significant improvements in performance. For example, we have already seen that even simple baselines, such as predicting a constant per-pixel average-chroma, achieve competitive performance when constrained to individual classes/domains (compare first and second row of Table~\ref{tb:imagenet_benchmark}). We expect similar performance improvements to be applicable to our approach.}

\paragraph{\textbf{Analysis of Computational and Memory Cost} }

\difftxt{In order to better understand the computational cost associated with our method, we break it down into two parts: encoding cost and decoding cost. For encoding, the first step in our approach involves generating multiple hypotheses for each pixel. This step takes 0.2 seconds on a GPU for a 256x256 image. Then, to perform a lossy-compression of the oracle we run a superpixel segmentation, which takes about 1 second per setting. The performance of the algorithm improves by exploring several superpixel settings in order to identify what size of superpixels is best to use. Using 5 pre-defined settings would raise the total time devoted to segmentation to about 5 seconds. Segmentation during decoding instead takes always only 1 second, since we only have to generate the segmentation for one superpixel setting (stored in the file).}
	
	\difftxt{In comparison, the runtimes reported in He et al.~\cite{he2009unified} indicate that the method of Cheng et al. takes 9393 seconds on average while the runtime of He et al. varies a lot depending on the image but takes an average of 786 seconds per image.}

\difftxt{Another important aspect for the applicability of our approach is the size of the model. Our model takes $105$ MB of space in its current state and, as a result, it is better suited for usage on the server side. However, with new approaches like those of Han et al.~\cite{han2015deep} showing promising model compression (up to $35$x), we believe our work may be usable on mobile devices too. On average, our model produces a file size saving of $800$+ bytes on a $256 \times 256$ image and the savings increase with the size of the image.} \LT{With model compression, the memory cost of our framework would be quickly amortized after the rendering of only a few high-resolution images. Furthermore, we believe that our approach is particularly valuable in scenarios where transmission capacity is limited (e.g., live streaming, or video conferencing over poor connections). The models of He et al.~\cite{he2009unified} and Cheng et al.~\cite{cheng2007learning} are much smaller than ours but, due to their prohibitive encoding and decoding costs, these methods are not applicable to scenarios such as video conferencing and live streaming.}

\section{Future Work}	
	In this paper we explored low-cost colorization for image compression using deep learning methods. The key idea behind our approach is the use of a multi-branch architecture that models the multi-modality of colors exhibited by many objects in the real world. Such architecture is well suited to be used in image compression scenarios: it is sufficient to store for each pixel a branch indicator denoting the branch producing the best approximation to the color value. We refer to this encoding as the oracle. We further improve on this scheme by presenting algorithms that leverage the spatial coherence of the oracle output to compactly approximate it via image subdivision. We demonstrate that the resulting system outperforms traditional JPEG color coding by a large margin, producing colors that are nearly indistinguishable from the ground truth at the storage cost of just a few hundred bytes for high-resolution pictures. Although our system was not designed to address the problem of zero-cost colorization (i.e., producing a single color hypothesis given a grayscale input without additional information being stored), our experiments indicate that our zero-cost variant is competitive with the best systems in this field. 
		
	In the future we plan to investigate architectures that yield similar compression gains for images in the high-quality regime. We also intend to study reliable metrics for understanding the color reproduction problem.
\difftxt{Furthermore, we aim to study learning objectives that would lead us to produce good colorization results in a near zero-cost regime. One such learning objective is a loss that selects a single branch for the colorization of an entire image (instead of selecting a branch for each pixel). Finally, we are interested in using our colorization training procedure from unlabelled images either as an unsupervised pretraining mechanism or as an auxiliary training task for high-level image understanding methods.	}
	
\section{Acknowledgments}
This work was funded in part by NSF CAREER award IIS-0952943 and NSF award CNS-120552. We gratefully acknowledge NVIDIA for the donation of GPUs used for portions of this work. We thank Karim Ahmed for helpful discussions and we also thank Larsson et al.~\cite{larsonLearningRepresentations} for providing clarifications regarding the results of their approach.

\bibliography{mybibfile}

\newpage
\appendix
\section{Network Architectures for models trained} \label{App:AppendixA}

We now describe the network architectures that we used with our proposed zero-cost and low-cost colorization framework. In all the subsequent tables, \difftxt{a Block} indicates a combination of layers in the following order \difftxt{:} convolution, batch normalization, scaling, \difftxt{ReLU}, convolution, batch normalization, scaling, elementwise addition with the previous layer indicated in the column for ``Residual Connection'' and a \difftxt{ReLU} for CIFAR100 architectures. \difftxt{A Block} in the \difftxt{ImageNet} architecture refers to blocks as defined in the Resnet-50 architecture from He et al.~\cite{deepresidualHe2015}. ``Block(2-9)'' \difftxt{indicates} that blocks $2$ to $9$ have the same \LT{number of} parameters as indicated in the row. The numbers in the filter column following \difftxt{a Block} indicate the number of filters used in each of the convolution layers for the block. We use \difftxt{zero-padding (referred to as Identity Projections by He et al.~\cite{deepresidualHe2015})} to match the number of layers in order to perform element-wise addition when we have to connect layers with different number of output channels by a residual connection. The second number in the projection column indicates the stride used during the projection. \difftxt{``Prev"} refers to a residual connection from the previous block in a chain of Blocks. Branch-$K$ refers to the $K$ branched module and its filters represent the number of \LT{kernels} in the convolution layers for each branch. 

	For all architectures we used as optimization the ADAM method proposed by Kingma and Ba.~\cite{kingma2014adam} with $\beta_1=0.9$ and $\beta_2=0.999$. We initialize all layers with the MSRA method for initialization as described by He et al.~\cite{he2015delving}. The colorization loss for training is described in equation~\ref{eq:branchLoss}.

	Table~\ref{tb:archCIFAR} shows the architecture of the model that was used to study how to train models with branching on the CIFAR100 dataset. The model was trained using the entire training set for CIFAR100 with mirroring for data augmentation with no cropping.
	
\paragraph{\bf Training Details for CIFAR100 Architecture}

	The model was trained for $40,000$ iterations with a mini-batch size of $64$ examples per GPU using 2 GPUs. We start training with a learning rate of $0.001$ and drop the learning rate twice during training by a factor of $10$ (after $10,000$ iterations and $20,000$ iterations). For experiments on the role of learning from scratch vs fine tuning we used the network pre-trained on the image categorization task by Ahmad et al.~\cite{nofe}.  For zero-cost colorization our branch prediction module uses the feature maps from the first layer of each branch (16 feature maps each). The branch predictor produces a per-pixel output of $5$ channels. This is followed by a \difftxt{Softmax} layer. 

\begin{table*}[ht!]  
  \centering
   \caption{The architecture of our deep network for colorization of images on the CIFAR100 dataset.\difftxt{BN stands for Batch Normalization, Sc stands for Scaling, RC stands for Residual Connection, Proj. stands for Projection. Y$^2$ means both  layers within \difftxt{a Block} have a certain component and N$^2$ means both  layers within \difftxt{a Block} do not have the component.}} \label{tb:archCIFAR}
\begin{tabular}{lcccccccc} 
%    \cmidrule{1-2}
    {\bf Layers} & Filters & Kernel & Stride & \difftxt{BN} & \difftxt{Sc} & \difftxt{ReLU} & \difftxt{RC} & \difftxt{Proj.}  \\
	\midrule
	Conv	-1				& 64 & 3 & 1 & Y & Y & Y & N\\
	Block-1			& 64-64 & 3-3 & 1-1 & \difftxt{Y$^2$} & \difftxt{Y$^2$} &  Y-N & Conv-1 & N\\
	Block(2-9) 		& 64-64 & 3-3 & 1-1 & \difftxt{Y$^2$} & \difftxt{Y$^2$} &  Y-N & \difftxt{Prev} & N\\
			
	Block-10			& 128-128 & 3-3 & 2-1 & \difftxt{Y$^2$} & \difftxt{Y$^2$} &  Y-N & Block-9 & 128-2\\						
	Block(11-18)			& 128-128 & 3-3 & 1-1 & \difftxt{Y$^2$} & \difftxt{Y$^2$} &  Y-N & \difftxt{Prev} & N\\	
	Block-19			& 256-256 & 3-3 & 2-1 & \difftxt{Y$^2$} & \difftxt{Y$^2$} &  Y-N & Block-18 & 256-2\\
	Block(20-29)			& 256-256 & 3-3 & 1-1 & \difftxt{Y$^2$} & \difftxt{Y$^2$} &  Y-N & \difftxt{Prev} & N\\
	UpSample-1			& 256 & 4 & 2 & N & N &  N & N & N\\		
	Block-30				& 128 & 3 & 1 & Y & Y &  Y & Block-18 & N\\		
	UpSample-2			& 128 & 4 & 2 & N & N &  N & N & N\\	
	Block-19				& 64 & 1 & 1 & Y & Y &  N & Block-9 & N\\				
	Branch-5				& (16,2) Each & \difftxt{3-3} & 1-1 & \difftxt{N$^2$} & \difftxt{N$^2$} & Y-N & \difftxt{N$^2$} & N\\
	Colorization \difftxt{loss} \\ 
	BranchPred & 5 & 3 & 1 & N & N & N & N & N \\
	Softmax \difftxt{loss} \\
    \midrule
  \end{tabular}
\end{table*}		

	Table~\ref{tb:archIN} describes the architecture used for colorization in our framework for the \difftxt{ImageNet} experiments. 
	
	\paragraph{\bf Training Details for \difftxt{ImageNet} Architecture}
	
	The model was trained for $120,000$ iterations with a mini-batch size of $20$ examples per GPU with 4 GPUs. We start with a learning rate of $0.0005$ and drop the learning rate twice during training by dividing it by $5$ (after $40,000$ iterations and $80,000$ iterations). We use a regularization of $0.0001$ for all layers except for the output layers in the branches and the output of the branch prediction module. For the Branch Prediction module in the \difftxt{ImageNet} experiments we used the features from the layers ``UpSample-4'', ``Conv-1'' and ``Block-3''. The feature maps from ``Conv-1'' and ``Block-3'' were upsampled \difftxt{independently} to match the resolution of the input grayscale image for dense prediction. The branch prediction module consisted of a convolution layer with $256$ filters (kernel size 3) followed by batch normalization, scaling and a second convolution layer with $64$ filters followed by batch normalization, scaling and a \difftxt{ReLU}. This was followed by a convolution layer with $5$ output channels per pixel and a Softmax layer. \LT{For experiments involving fine-tuning of the base model we replicated the input grayscale image $3$ times \difftxt{to} form a $3$ channel input for the network.}

\begin{table*}[ht]
  \centering
  \caption{The architecture of our deep network for colorization of images on \difftxt{ImageNet}.\difftxt{BN stands for Batch Normalization, Sc stands for Scaling, RC stands for Residual Connection, Proj. stands for Projection. Y$^3$ means all the three layers within \difftxt{a Block} have a certain component and Y$^2$N means the first two layers within \difftxt{a Block} have a component but the third layer does not.}} 
\begin{tabular}{lcccccccc} \label{tb:archIN}
%    \cmidrule{1-2}
    {\bf Layers} & Filters & Kernel & Stride & \difftxt{BN} & \difftxt{Sc} & \difftxt{ReLU} & \difftxt{RC} & \difftxt{Proj.}  \\
	\midrule
	Conv-1				& 64 & 7 & 2 & Y & Y & Y & N\\
	Pool-1 				&    & 3 & 2 &   &   &   &  \\
	Block-1			& 64-64-256 & 1-3-1 & 1-1-1 & \difftxt{Y$^3$} & \difftxt{Y$^3$} &  \difftxt{Y$^2$N} & Pool-1 & 256-1\\
	Block(2-3)			& 64-64-256 & 1-3-1 & 1-1-1 & \difftxt{Y$^3$} & \difftxt{Y$^3$} &  \difftxt{Y$^2$N} & \difftxt{Prev} & N\\	
	Block-4			& 128-128-512 & 1-3-1 & 2-1-1 & \difftxt{Y$^3$} & \difftxt{Y$^3$} &  \difftxt{Y$^2$N} & Block-3 & 512-2\\
	Block(5-7)			& 128-128-512 & 1-3-1 & 1-1-1 & \difftxt{Y$^3$} & \difftxt{Y$^3$} &  \difftxt{Y$^2$N} & \difftxt{Prev} & N\\
	Block-8			& 256-256-1024 & 1-3-1 & 2-1-1 & \difftxt{Y$^3$} & \difftxt{Y$^3$} &  \difftxt{Y$^2$N} & Block-7 & 1024-2\\
	Block(9-13)			& 256-256-1024 & 1-3-1 & 1-1-1 & \difftxt{Y$^3$} & \difftxt{Y$^3$} &  \difftxt{Y$^2$N} & \difftxt{Prev} & N\\	
	Block-14			& 512-512-2048 & 1-3-1 & 2-1-1 & \difftxt{Y$^3$} & \difftxt{Y$^3$} &  \difftxt{Y$^2$N} & Block-13 & 2048-2\\	
	Block(15-16)			& 512-512-2048 & 1-3-1 & 1-1-1 & \difftxt{Y$^3$}& \difftxt{Y$^3$} &  \difftxt{Y$^2$N} & \difftxt{Prev} & N\\		
	Block-17				& 1024 & 1 & 1 & Y & Y &  N & N & N\\		
	UpSample-1			& 1024 & 4 & 2 & N & N &  N & Block-13 & N\\		
	Block-18				& 512 & 1 & 1 & Y & Y &  N & N & N\\			
	UpSample-2			& 512 & 4 & 2 & N & N &  N & Block-7 & N\\	
	Block-19				& 256 & 1 & 1 & Y & Y &  N & N & N\\			
	UpSample-3			& 256 & 4 & 2 & N & N &  N & Block-3 & N\\	
	Block-20				& 64 & 1 & 1 & Y & Y &  N & N & N\\			
	UpSample-3			& 64 & 4 & 2 & N & N &  N & Conv-1 & N\\
	Block-21				& 32 & 1 & 1 & Y & Y &  N & N & N\\			
	UpSample-4			& 32 & 4 & 2 & N & N &  N & N & N\\
	Branch-5				& 2-Each & 3-Each & \difftxt{1} & N & N & N & N & N\\
	Colorization \difftxt{loss} \\
	BranchPred & 5 & 3 & 1 & N & N & N & N & N \\
	Softmax \difftxt{loss} \\
    \midrule
  \end{tabular}
\end{table*}	

\paragraph{\difftxt{\bf Additional comparisons with JPEG}}
We now proceed to shown additional results comparing our approach with JPEG. Whereas our approach occasionally misses out small details that JPEG is able to capture, overall our approach produces colors that are more correct and show less visible artifacts at file sizes that are an order of magnitude smaller than JPEG.

\begin{table*}[t!]\label{tb:jpegMore1}
\centering
\setlength{\tabcolsep}{1pt}
\begin{tabular}{c|c|ccc}
\toprule
 Image & JPEG~\cite{jpeg} & \multicolumn{3}{|c}{Our approach} \\ 
\hline
 & $1,263$ bytes & $99$ bytes & $178$ bytes & $373$ bytes\\
\includegraphics[width=3.0cm]{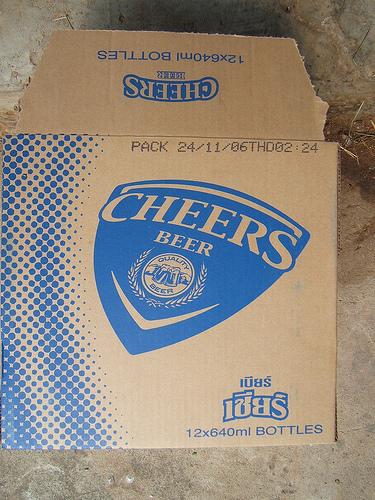}&\includegraphics[width=3.0cm]{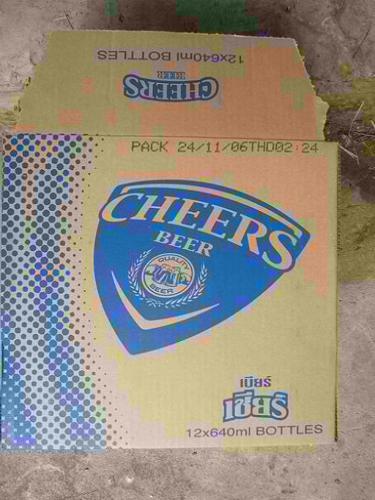}&\includegraphics[width=3.0cm]{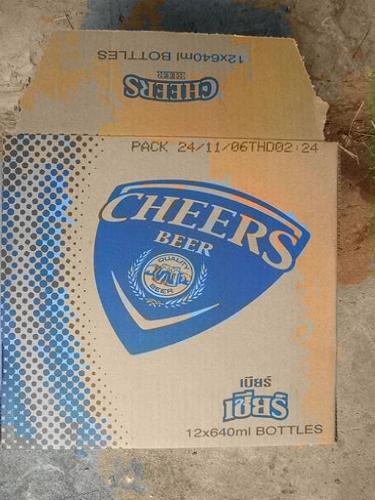}&\includegraphics[width=3.0cm]{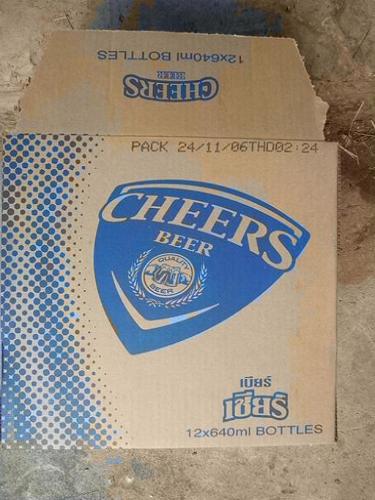}&\includegraphics[width=3.0cm]{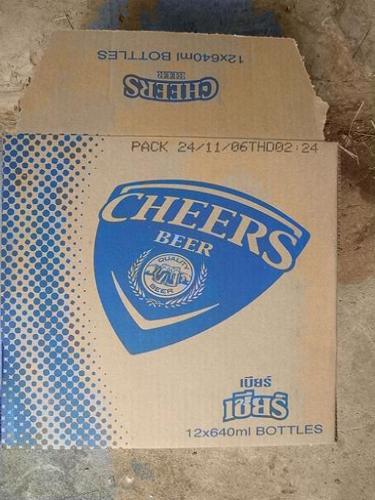}\\ 
\hline
 & $1,132$ bytes & $99$ bytes & $399$ bytes & $468$ bytes\\
\includegraphics[width=3.0cm]{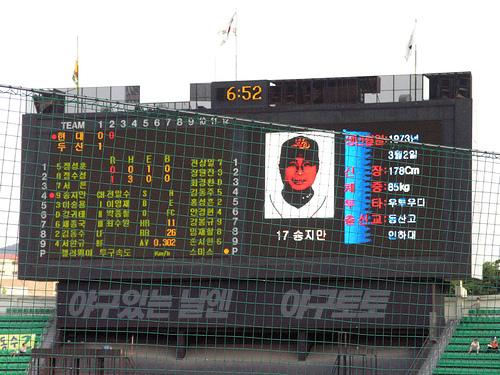}&\includegraphics[width=3.0cm]{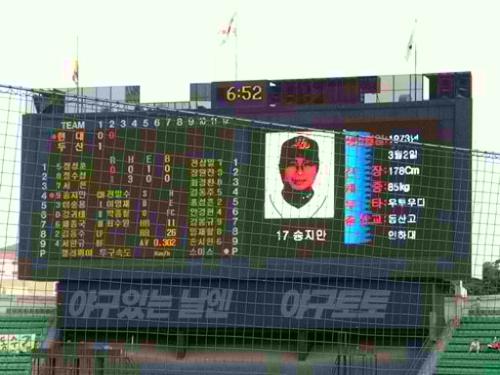}&\includegraphics[width=3.0cm]{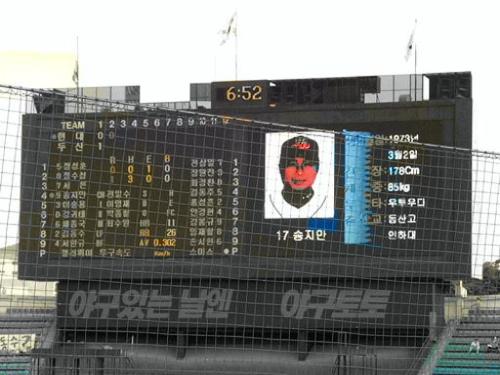}&\includegraphics[width=3.0cm]{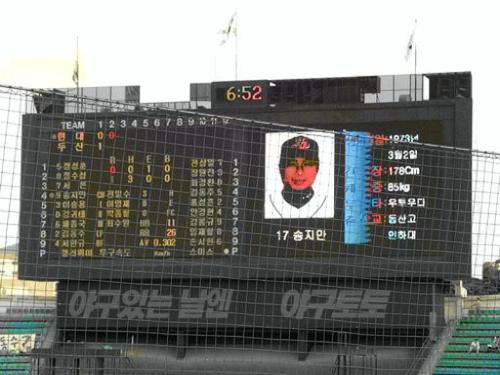}&\includegraphics[width=3.0cm]{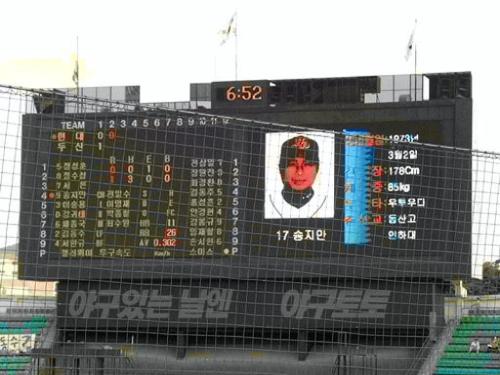}\\ 
\hline
 & $1,160$ bytes & $98$ bytes & $242$ bytes & $470$ bytes\\
\includegraphics[width=3.0cm]{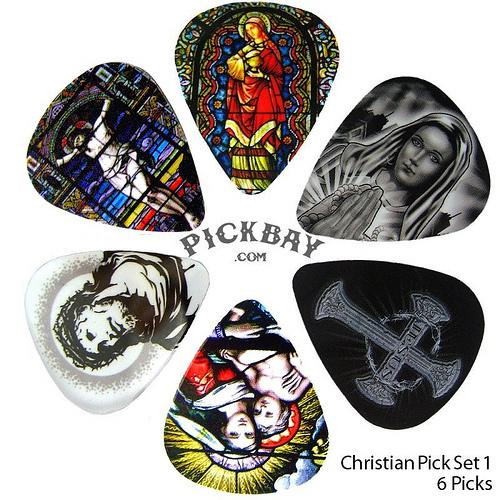}&\includegraphics[width=3.0cm]{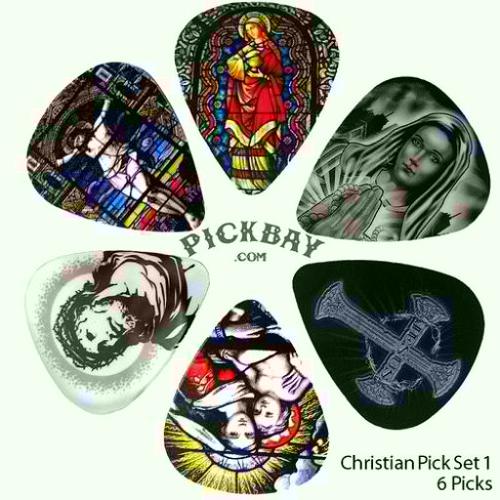}&\includegraphics[width=3.0cm]{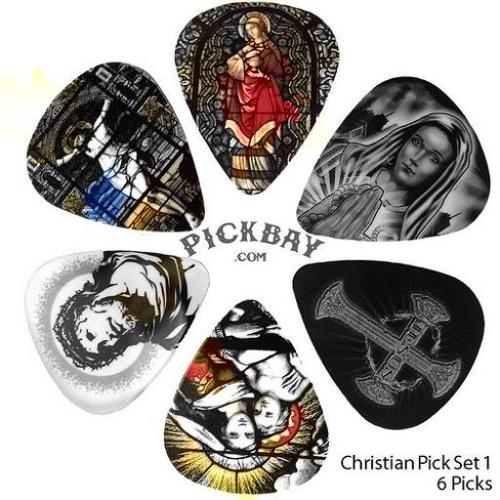}&\includegraphics[width=3.0cm]{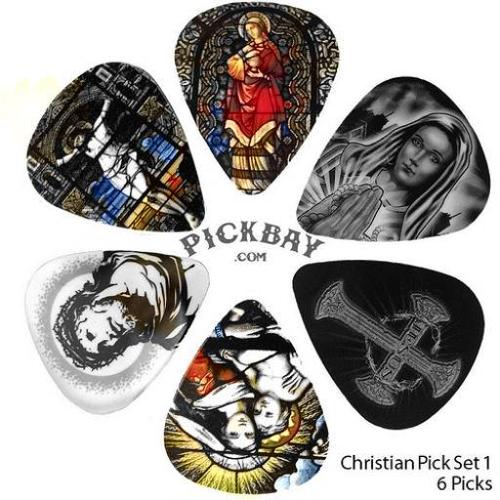}&\includegraphics[width=3.0cm]{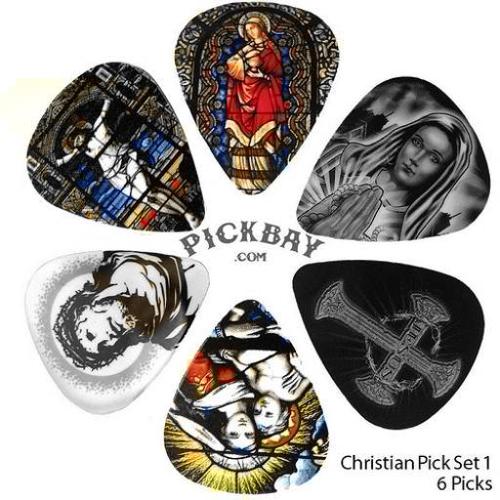}\\ 
\hline
 & $1,122$ bytes & $96$ bytes & $234$ bytes & $351$ bytes\\
\includegraphics[width=3.0cm]{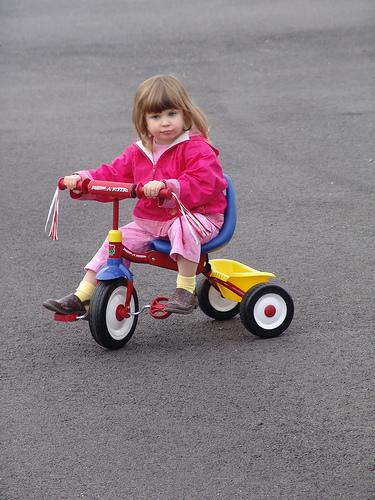}&\includegraphics[width=3.0cm]{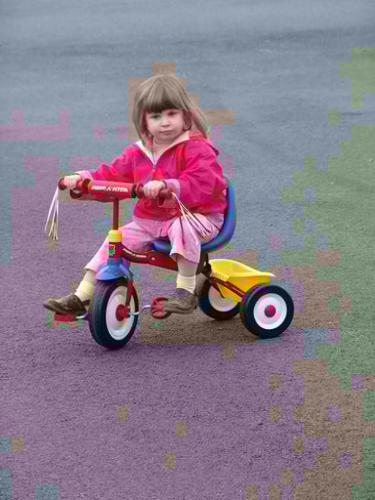}&\includegraphics[width=3.0cm]{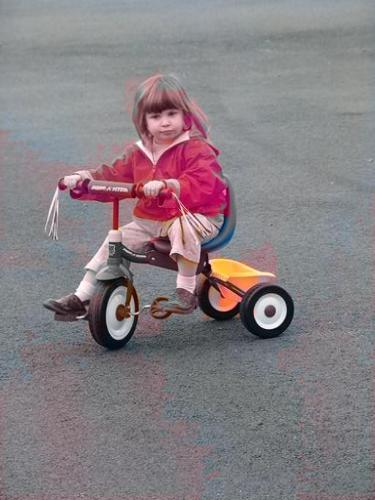}&\includegraphics[width=3.0cm]{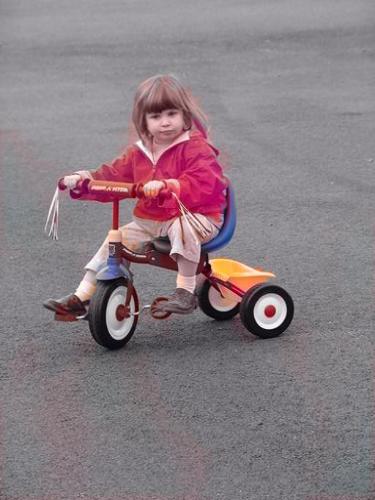}&\includegraphics[width=3.0cm]{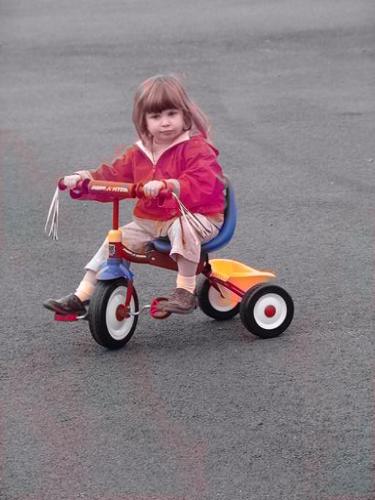}\\ 
\hline
 & $1,164$ bytes & $96$ bytes & $234$ bytes & $351$ bytes\\
\includegraphics[width=3.0cm]{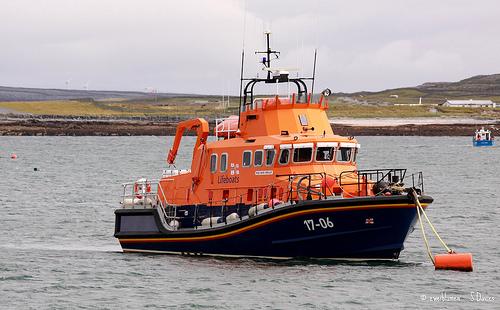}&\includegraphics[width=3.0cm]{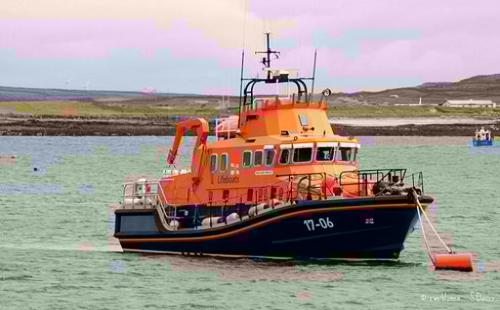}&\includegraphics[width=3.0cm]{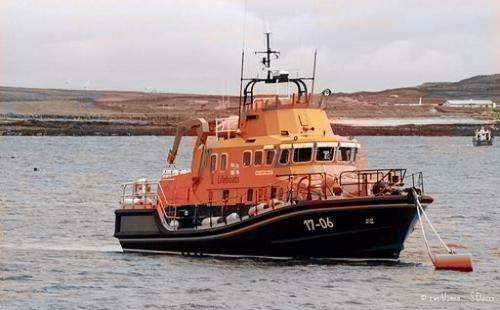}&\includegraphics[width=3.0cm]{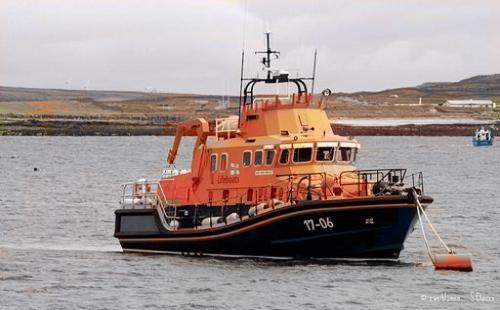}&\includegraphics[width=3.0cm]{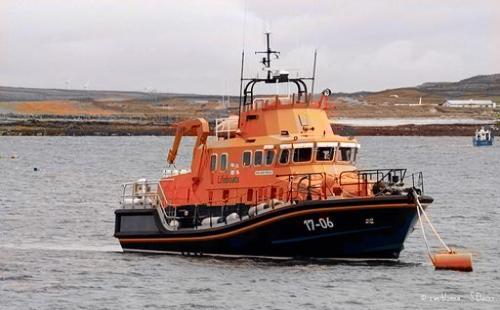}\\ 
\hline
 & $1,189$ bytes & $97$ bytes & $254$ bytes & $487$ bytes\\
\includegraphics[width=3.0cm]{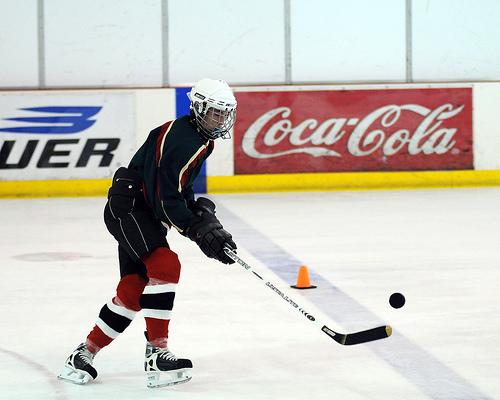}&\includegraphics[width=3.0cm]{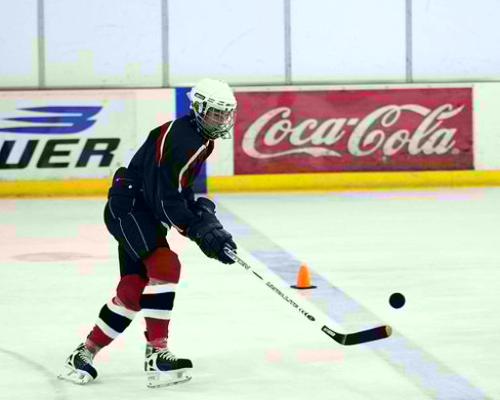}&\includegraphics[width=3.0cm]{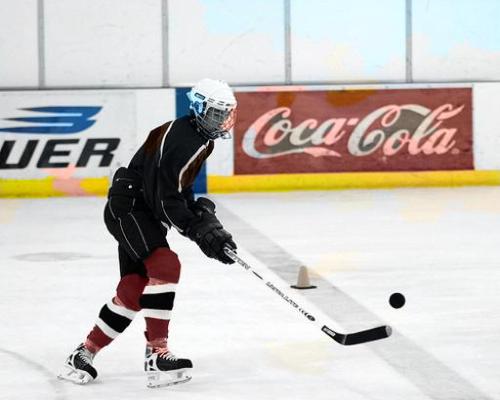}&\includegraphics[width=3.0cm]{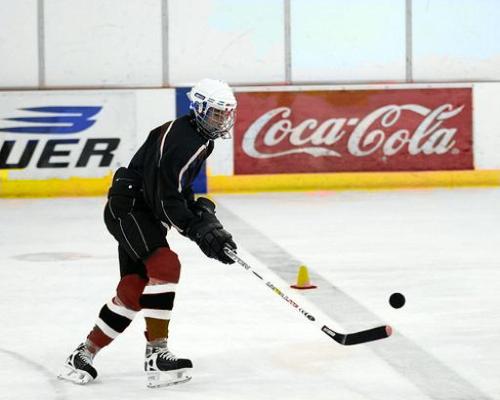}&\includegraphics[width=3.0cm]{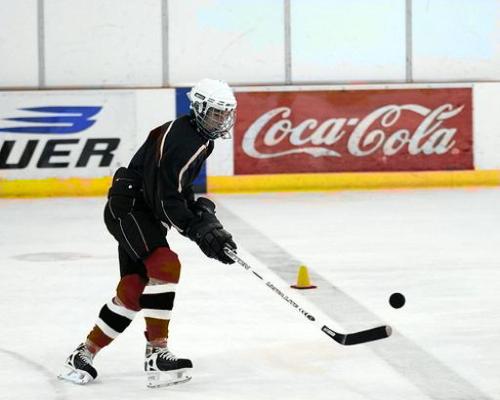}\\ 
\hline
\end{tabular}
\caption{We show visualizations generated by our proposed low-cost framework and JPEG color coding~\cite{jpeg} (for JPEG we report the storage space required to compress {\em only} the color channels). Our approach, produces vibrant realistic looking images at only about 1/$6^{th}$ of the storage size required by JPEG.}
\end{table*}

\end{document}